\documentclass{colt2015} 

\usepackage{algorithm}
\usepackage{algorithmic}
\usepackage{bm}
\usepackage{amsfonts}
\usepackage{wrapfig}
\usepackage{graphicx}

\newlength{\subfigwidth}
\newlength{\subfigcolsep}
\title[Regret Lower Bound and Optimal Algorithm in Dueling Bandit Problem]{Regret Lower Bound and Optimal Algorithm\\ in Dueling Bandit 	Problem}
\usepackage{times}
 \coltauthor{\Name{Junpei Komiyama} \Email{junpei@komiyama.info}\\
 \addr The University of Tokyo\\
 \Name{Junya Honda} \Email{honda@stat.t.u-tokyo.ac.jp}\\
 \addr The University of Tokyo\\
 \Name{Hisashi Kashima} \Email{kashima@i.kyoto-u.ac.jp}\\
 \addr Kyoto University\\
 \Name{Hiroshi Nakagawa} \Email{nakagawa@dl.itc.u-tokyo.ac.jp}\\
 \addr The University of Tokyo\\
 }

\DeclareMathOperator*{\argmin}{\mathop{\rm arg~min}}
\newtheorem{fact}[theorem]{Fact}

\newcommand{\Prob}{\mathbb{P}}
\newcommand{\Expect}{\mathbb{E}}
\newcommand{\Indicator}{{\mathbf{1}}}
\newcommand{\Ind}{\Indicator}

\newcommand{\hatDeltaij}[2]{\hat{\Delta}_{#1,#2}}

\newcommand{\Suf}{\mathrm{Suf}}
\newcommand{\hSetO}[1]{{\hat{\mathcal{O}}_{#1}}}
\newcommand{\SetO}[1]{{\mathcal{O}_{#1}}}
\newcommand{\SetOd}[1]{{\mathcal{O}'_{#1}}}

\newcommand{\Regret}{{R}}

\newcommand{\Bernoulli}{\mathrm{Bernoulli}}
\newcommand{\muij}[2]{{\mu_{{#1},{#2}}}}
\newcommand{\muijd}[2]{{\mu_{{#1},{#2}}'}}
\newcommand{\hatmut}[2]{{\hat{\mu}}_{{#1},{#2}}(t)}
\newcommand{\hatmun}[3]{{\hat{\mu}}_{{#1},{#2}}^{#3}}
\newcommand{\Xij}{{\hat{X}_{i,j}}}
\newcommand{\myX}[2]{{\hat{X}_{{#1},{#2}}}}
\newcommand{\myNt}[2]{{N_{{#1},{#2}}(t)}}
\newcommand{\myNtt}[3]{{N_{{#1},{#2}}(#3)}}
\newcommand{\myNT}[2]{{N_{{#1},{#2}}(T)}}
\newcommand{\myNnon}[2]{{N_{{#1},{#2}}}}
\newcommand{\myNbeat}[2]{{N_{{#1}>{#2}}(t)}}
\newcommand{\Mat}{M}
\newcommand{\EJ}{\mathcal{J}}
\newcommand{\EU}{\mathcal{U}}
\newcommand{\EA}{\mathcal{Y}}
\newcommand{\EB}{\mathcal{Z}}

\newcommand{\ED}{\mathcal{D}}
\newcommand{\EE}{\mathcal{E}}

\newcommand{\ist}{{i^*(t)}}
\newcommand{\Ist}{{I^*(t)}}

\newcommand{\nonwinner}{{[K] \setminus \{1\}}}
\newcommand{\nn}{\nonumber\\}
\newcommand{\nsuf}[2]{{N_{{#1},{#2}}^{\Suf}(\mysmalldelta)}}
\newcommand{\myDelta}[2]{{\Delta_{{#1},{#2}}}}

\newcommand{\bsi}{{b^{\star}(i)}}
\newcommand{\bsiarg}[1]{{b^{\star}(#1)}}
\newcommand{\hatbs}[1]{{\hat{b}^{\star}(#1)}}
\newcommand{\hatbsi}{{\hatbs{i}}}

\newcommand{\dsuf}{{\Delta_i^{\text{suf}}}}

\newcommand{\hKL}{\mathrm{\widehat{KL}}}
\newcommand{\taut}[1]{{\tau(#1)}}

\newcommand{\rd}{\mathrm{d}}
\newcommand{\e}{\mathrm{e}}
\newcommand{\Natural}{\mathbb{N}}

\newcommand{\Real}{\mathbb{R}}
\newcommand{\Tinit}{T_{\mathrm{init}}}

\newcommand{\CoefA}{A}
\newcommand{\mysmalldelta}{{\delta}}
\newcommand{\mysmallepsilon}{{\epsilon}}

\newenvironment{mynotitleproof}[1]{}{\hfill\BlackBox\\[2mm]}
\newcommand{\myalpha}{\alpha}
\newcommand{\morder}{\mathcal{M}_{\mathrm{o}}}
\newcommand{\mcondor}{\mathcal{M}_{\mathrm{C}}}
\newcommand{\model}{\mathcal{M}}
\newcommand{\qed}{\hfill\BlackBox\\[2mm]}

\begin{document}

\maketitle

\begin{abstract}
We study the $K$-armed dueling bandit problem, a variation of the standard stochastic bandit problem
 where the feedback is limited to relative comparisons of a pair of arms.
We introduce a tight asymptotic regret lower bound that is based on the information divergence.
An algorithm that is inspired by the Deterministic Minimum Empirical Divergence algorithm (Honda and Takemura, 2010) is proposed, and its regret is analyzed.
The proposed algorithm is found to be the first one with a regret upper bound that matches the lower bound.
Experimental comparisons of dueling bandit algorithms show that the proposed algorithm significantly outperforms existing ones.
\end{abstract}


\section{Introduction}

A multi-armed bandit problem is a crystallized instance of a sequential decision-making problem in an uncertain environment, and it can model many real-world scenarios.
This problem involves conceptual entities called arms,
 and a forecaster who tries to identify good arms from bad ones.
At each round, the forecaster draws one of the $K$ arms and receives a corresponding reward.
The aim of the forecaster is to maximize the cumulative reward over rounds, which is achieved by running an algorithm that balances  the exploration (acquisition of information) and the exploitation (utilization of information).

While it is desirable to obtain direct feedback from an arm, in some cases such direct feedback is not available.
In this paper, we consider a version of the standard stochastic bandit problem called the $K$-armed dueling bandit problem \citep{DBLP:conf/colt/YueBKJ09},
 in which the forecaster receives relative feedback, which specifies which of two arms is preferred.
Although the original motivation of the dueling bandit problem arose in the field of information retrieval,
 learning under relative feedback is universal to many fields, such as recommender systems \citep{Gemmis09preferencelearning}, graphical design \citep{DBLP:conf/sca/BrochuBF10}, and natural language processing \citep{DBLP:conf/acl/ZaidanC11}, which involve explicit or implicit feedback provided by humans. 

\noindent\textbf{Related work:}
Here, we briefly discuss the literature of the $K$-armed dueling bandit problem.
The problem involves a preference matrix $\Mat =\{\muij{i}{j}\} \in \Real^{K \times K}$, whose $ij$ entry $\muij{i}{j}$ corresponds to the probability that arm $i$ is preferred to arm $j$.

Most algorithms assume that the preference matrix has certain properties.
Interleaved Filter (IF) \citep{DBLP:journals/jcss/YueBKJ12} and Beat the Mean Bandit (BTM) \citep{DBLP:conf/icml/YueJ11}, early algorithms proposed for solving the dueling bandit problem, require the arms to be totally ordered, that is, $i \succ j\Leftrightarrow \muij{i}{j} > 1/2$.
Moreover, IF assumes \textit{stochastic transitivity}: for any triple $(i,j,k)$ with $i \succ j \succ k$, $\muij{i}{k} \geq \max{\{\muij{i}{j}, \muij{j}{k}\}}$. Unfortunately, stochastic transitivity does not hold in many real-world settings \citep{DBLP:conf/icml/YueJ11}.
BTM relaxes this assumption by introducing \textit{relaxed stochastic transitivity}: there exists $\gamma \geq 1$ such that for all pairs $(j,k)$ with $1 \succ j \succ k$, $\gamma \muij{1}{k} \geq \max{\{\muij{1}{j}, \muij{j}{k}\}}$ holds. The drawback of BTM is that it requires the explicit value of $\gamma$ on which the performance of the algorithm depends.
\cite{DBLP:conf/icml/UrvoyCFN13} considered a wide class of sequential learning problems with bandit feedback that includes the dueling bandit problem. They proposed the Sensitivity Analysis of VAriables for Generic Exploration (SAVAGE) algorithm, which empirically outperforms IF and BTM for moderate $K$.
Among the several versions of SAVAGE, the one called Condorcet SAVAGE makes the \textit{Condorcet assumption} and performed the best in their experiment. The Condorcet assumption is that there is a unique arm that is superior to the others. Unlike the two transitivity assumptions, the Condorcet assumption does not require the arms to be totally ordered and is less restrictive.
IF, BTM, and SAVAGE either explicitly require the number of rounds $T$, or implicitly require $T$ to determine the confidence level $\delta$.

Recently, an algorithm called Relative Upper Confidence Bound (RUCB) \citep{DBLP:conf/icml/ZoghiWMR14} was proven to have an $O(K \log{T})$ regret bound under the Condorcet assumption.
 RUCB is based on the upper confidence bound index \citep{LaiRobbins1985,Agr95,auerfinite} that is widely used in the field of bandit problems. RUCB is \textit{horizonless}: it does not require $T$ beforehand and runs for any duration.
\cite{zoghiwsdm2015} extended RUCB into the mergeRUCB algorithm under the Condorcet assumption as well as the assumption that a portion of the preference matrix is informative (i.e., different from $1/2$). They reported that mergeRUCB outperformed RUCB when $K$ was large.
\cite{DBLP:conf/icml/AilonKJ14} proposed three algorithms named Doubler, MultiSBM, and Sparring. MultiSBM is endowed with an $O(K \log{T})$ regret bound and Sparring was reported to outperform IF and BTM in their simulation.
These algorithms assume that the pairwise feedback is generated from the non-observable utilities of the selected arms. 
The existence of the utility distributions associated with individual arms restricts the structure of the preference matrix.

In summary,
most algorithms either has $O(K^2\log T)$ regret under the Condorcet assumption 
(SAVAGE) or 
require additional assumptions
to achieve $O(K\log T)$ regret (IF, BTM, MultiSBM, and mergeRUCB).
To the best of our knowledge, RUCB is the only algorithm with an $O(K\log T)$ regret bound\footnote{\cite{DBLP:journals/corr/ZoghiWMR13} first proposed RUCB with an $O(K^2 \log T)$ regret bound and later modified it by adding a randomization procedure to assure $O(K \log{T}$) regret in \cite{DBLP:conf/icml/ZoghiWMR14}.}.
The main difficulty of the dueling bandit problem lies in that, 
there are $K-1$ candidates of actions to test ``how good'' each arm $i$ is.
A naive use of the confidence bound requires every pair of arms to be compared $O(\log{T})$ times and yields an $O(K^2 \log{T})$ regret bound.

\noindent\textbf{Contribution:}
In this paper, we propose an algorithm called Relative Minimum Empirical Divergence (RMED). This paper contributes to our understanding of the dueling bandit problem in the following three respects.
\begin{itemize}
\vspace{-0.6em}
 \item \textbf{The regret lower bound:} Some studies (e.g., \citealp{DBLP:journals/jcss/YueBKJ12}) have shown that the $K$-armed dueling bandit problem has a $\Omega(K \log{T})$ regret lower bound. In this paper, we further analyze this lower bound to obtain the optimal constant factor for models satisfying the Condorcet assumption.
Furthermore, we show that the lower bound is the same under the total order assumption.
This means that optimal algorithms under the
Condorcet assumption also achieve a lower bound of regret under the total order assumption
even though such algorithms do not know that the arms are totally ordered.
\vspace{-0.6em}
 \item \textbf{An optimal algorithm:} The regret of RMED is not only $O(K \log{T})$, but also optimal in the sense that its constant factor matches the asymptotic lower bound under the Condorcet assumption. RMED is the first optimal algorithm in the study of the dueling bandit problem.
\vspace{-0.6em}
 \item \textbf{Empirical performance assessment:} The performance of RMED is extensively evaluated by using five datasets: two synthetic datasets, one including preference data, and two including ranker evaluations in the information retrieval domain.
\vspace{-0.5em}
\end{itemize}

\section{Problem Setup}
\label{sec:setup}

The $K$-armed dueling bandit problem involves $K$ arms that are indexed as $[K] = \{1,2,\dots,K\}$. Let $\Mat \in \Real^{K \times K}$ be a preference matrix whose $ij$ entry $\muij{i}{j}$ corresponds to the probability that arm $i$ is preferred to arm $j$.
At each round $t=1,2,\dots,T$, the forecaster selects a pair of arms $(l(t), m(t)) \in [K]^2$, then receives a relative feedback $\myX{l(t)}{m(t)}(t) \sim \Bernoulli(\muij{l(t)}{m(t)})$ that indicates which of $(l(t), m(t))$ is preferred. By definition, $\muij{i}{j} = 1 - \muij{j}{i}$ holds for any $i,j \in [K]$ and $\muij{i}{i} = 1/2$. 

Let $\myNt{i}{j}$ be the number of comparisons of pair $(i,j)$ and $\hatmut{i}{j}$ be the empirical estimate of $\muij{i}{j}$ at round $t$. In building statistics by using the feedback, we treat pairs without taking their order into consideration. Therefore, for $i \neq j $, $\myNt{i}{j} = \sum_{t'=1}^{t-1} (\Ind\{l(t')=i, m(t')=j\} + \Ind\{l(t')=j, m(t')=i\})$ and $\muij{i}{j} = (\sum_{t'=1}^{t-1} (\Ind\{l(t')=i, m(t')=j, \myX{l(t')}{m(t')}(t')=1\} + \Ind\{l(t')=j, m(t')=i, \myX{l(t')}{m(t')}(t')=0\}))/\myNt{i}{j}$, where $\Ind[\cdot]$ is the indicator function.
For $j \neq i$, let $\myNbeat{i}{j}$ be the number of times $i$ is preferred over $j$. Then, $\hatmut{i}{j} = \myNbeat{i}{j}/\myNt{i}{j}$, where we set $0/0 = 1/2$ here. Let $\hatmut{i}{i}=1/2$.

Throughout this paper, we will assume that the preference matrix has
a Condorcet winner \citep{DBLP:conf/icml/UrvoyCFN13}.
Here we call an arm $i$ the Condorcet winner
if $\muij{i}{j} >1/2$ for any $j\in [K]\setminus \{i\}$.
Without loss of generality, we will assume that arm $1$ is the Condorcet winner.
The set of preference matrices which have a Condorcet winner is denoted
by $\mcondor$.
We also define the set of preference matrices satisfying the total order
by $\morder\subset \mcondor$; that is,
the relation $i\prec j\Leftrightarrow \muij{i}{j} < 1/2$ induces a total order
iff $\{ \muij{i}{j} \}\in\morder$.

Let $\myDelta{i}{j} = \muij{i}{j} -1/2$.
We define the regret per round as $r(t) = (\myDelta{1}{i}+\myDelta{1}{j})/2$ when the pair $(i,j)$ is compared.
The expectation of the cumulative regret,
$ \Expect[ \Regret(T) ] = \Expect\left[ \sum_{t=1}^T r(t) \right]
$is used to measure the performance of an algorithm.
The regret increases at each round unless the selected pair is $(l(t),m(t)) = (1,1)$.

\subsection{Regret lower bound in the $K$-armed dueling bandits}
\label{subsec:duelinglower}

In this section we provide an asymptotic regret lower bound when $T \rightarrow \infty$.
Let the \text{superiors} of arm $i$ be a set $\SetO{i} = \{j|j \in [K], \muij{i}{j} <1/2\}$, that is, the set of arms that is preferred to $i$ on average.
The essence of the $K$-armed dueling bandit problem is how to eliminate each arm $i \in \nonwinner$ by making sure that arm $i$ is not the Condorcet winner. To do so, the algorithm uses some of the arms in $\SetO{i}$ and compares $i$ with them.

A dueling bandit algorithm is strongly consistent for model $\model \subset \mcondor$
iff it has $\Expect[\Regret(T)] = o(T^a)$
regret for any $a > 0$ and any $\Mat \in \model$.
The following lemma is on the number of comparisons of suboptimal arm pairs.
\begin{lemma} {\rm (The lower bound on the number of suboptimal arm draws)}
(i)
Let an arm $i \in \nonwinner$ and
preference matrix $\Mat \in \mcondor$ be arbitrary.
Given any strongly consistent algorithm for model $\mcondor$, we have
\begin{equation}
 \Expect\left\{\sum_{j \in \SetO{i}} d(\muij{i}{j}, 1/2) N_{i,j}(T) \right\} \geq (1 - o(1)) \log{T}, \label{ineq:drawlower}
\end{equation}
where $d(p,q) = p \log{\frac{p}{q}} + (1-p)\log{\frac{1-p}{1-q}}$ is the KL divergence between two Bernoulli distributions with  parameters $p$ and $q$.
(ii) Furthermore, inequality \eqref{ineq:drawlower} holds
for any $\Mat \in \morder$ given any strongly consistent algorithm for $\morder$.
\label{lem:drawlower}
\end{lemma}
Lemma \ref{lem:drawlower} states that, for arbitrary arm $j \in \SetO{i}$, an algorithm needs to make $\log{T}/d(\muij{i}{j}, 1/2)$ comparisons between arms $i$ and $j$ to be convinced that arm $i$ is inferior to arm $j$ and thus $i$ is not the Condorcet winner. Since the regret increase per round of comparing arm $i$ with $j$ is $(\myDelta{1}{i} + \myDelta{1}{j})/2$, eliminating arm $i$ by comparing it with $j$ incurs a regret of
\begin{equation}
 \frac{(\myDelta{1}{i} + \myDelta{1}{j})\log{T}}{2 d(\muij{i}{j}, 1/2)}. \label{regret_ij}
\end{equation}
Therefore, the total regret is bounded from below by comparing each arm $i$ with an arm $j$ that minimizes
\eqref{regret_ij} and the regret lower bound is formalized in the following theorem.
\begin{theorem} {\rm (The regret lower bound)}
(i) Let the preference matrix $\Mat \in \mcondor$ be arbitrary. For any strongly consistent algorithm for model $\mcondor$,
\begin{equation}
 \liminf_{T \rightarrow \infty} \frac{\Expect[\Regret(T)]}{\log{T}} \geq \sum_{i \in \nonwinner} \min_{j \in \SetO{i}} \frac{\myDelta{1}{i} + \myDelta{1}{j}}{2 d(\muij{i}{j}, 1/2)} \label{ineq:regretlower}
\end{equation}
holds.
(ii) Furthermore,
inequality \eqref{ineq:regretlower} holds for any $\Mat \in \morder$ given any strongly consistent algorithm for $\morder$.
\label{thm:regretlower}
\end{theorem}
The proof of Lemma \ref{lem:drawlower} and Theorem \ref{thm:regretlower} can be found in Appendix \ref{sec:lowerboundproof}.
The proof of Lemma \ref{lem:drawlower} is similar to that of \citet[Theorem 1]{LaiRobbins1985}
for the standard multi-armed bandit problem but differs in the following point that is characteristic
to the dueling bandit.
To achieve a small regret in the dueling bandit, it is necessary to compare the arm $i$ with itself if $i$ is the Condorcet winner.
However, we trivially know that $\muij{i}{i}=1/2$ without sampling
and such a comparison yields no information to distinguish possible preference matrices.
We can avoid this difficulty by evaluating $\myNnon{i}{j}$ and $\myNnon{i}{i}$ in different ways.

\section{RMED1 Algorithm}
\label{sec:rmedone}

In this section, we first introduce the notion of empirical divergence. Then, on the basis of the empirical divergence, we formulate the RMED1 algorithm.

\begin{algorithm}[t]
 \caption{Relative Minimum Empirical Divergence (RMED) Algorithm}
 \label{alg:rmedbase}
\begin{algorithmic}[1]
   \STATE \textbf{Input:} $K$ arms, $f(K) \ge 0$. $\myalpha > 0$ (RMED2FH, RMED2). $T$ (RMED2FH).
   \STATE $L \leftarrow \begin{cases}
    1 & \text{(RMED1, RMED2)} \\
    \lceil \myalpha \log{\log{T}} \rceil & \text{(RMED2FH)}
  \end{cases}$.
   \STATE \textbf{Initial phase:} draw each pair of arms $L$ times. At the end of this phase, $t= L (K-1)K/2$.
   \IF{RMED2FH}
     \STATE For each arm $i \in [K]$, fix $\hatbsi$ by \eqref{ineq:hatbsi}.
   \ENDIF
   \STATE $L_C, L_R \leftarrow [K], L_N \leftarrow \emptyset$.
   \WHILE{$t \le T$}
     \IF{RMED2}
       \STATE Draw all pairs $(i,j)$ until it reaches $\myNt{i}{j} \geq \myalpha \log{\log{t}}$. $t \leftarrow t+1$ for each draw. \label{line:rmed2exploration}
     \ENDIF
     \FOR{$l(t) \in L_C$ in an arbitrarily fixed order}
       \STATE Select $m(t)$ by using $\begin{cases}
    \text{Algorithm \ref{alg:rmedverone}} & \text{(RMED1)} \\
    \text{Algorithm \ref{alg:rmedverttwo}} & \text{(RMED2, RMED2FH)} \\
    \end{cases}$.
       \STATE Draw arm pair ($l(t)$, $m(t)$).
       \STATE $L_R \leftarrow L_R \setminus \{l(t)\}$.
       \STATE $L_N \leftarrow L_N \cup \{j\}$ (without a duplicate) for any $j \notin L_R$ such that $\EJ_j(t)$ holds.
       \STATE $t \leftarrow t+1$.
     \ENDFOR
     \STATE $L_C, L_R \leftarrow L_N$, $L_N \leftarrow \emptyset$.
   \ENDWHILE
\end{algorithmic}
\end{algorithm}

\begin{algorithm}[t]
 \caption{RMED1 subroutine for selecting $m(t)$}
 \label{alg:rmedverone}
\begin{algorithmic}[1]
  \STATE $\hSetO{l(t)}(t) \leftarrow \{j \in [K]\setminus\{l(t)\}|  \hatmut{l(t)}{j} \leq 1/2\}$
  \IF{$\ist \in \hSetO{l(t)}(t)$ or $\hSetO{l(t)}(t)=\emptyset$}
    \STATE $m(t) \leftarrow \ist$. \label{line:istselectedverone}
  \ELSE
    \STATE $m(t) \leftarrow \argmin_{j \neq l(t)} \hatmut{l(t)}{j}$.
  \ENDIF
\end{algorithmic}
\end{algorithm}

\subsection{Empirical divergence and likelihood function}

In inequality \eqref{ineq:drawlower} of Section \ref{subsec:duelinglower}, we have seen that $\sum_{j \in \SetO{i}} d(\muij{i}{j}, 1/2) \myNT{i}{j}$, the sum of the divergence between $\muij{i}{j}$ and $1/2$ multiplied by the number of comparisons between $i$ and $j$, is the characteristic value that defines the minimum number of comparisons.
The empirical estimate of this value is fundamentally useful for evaluating how unlikely arm $i$ is to be the Condorcet winner. 
Let the \text{opponents} of arm $i$ at round $t$ be the set $\hSetO{i}(t) = \{j| j \in [K] \setminus \{i\}, \hatmut{i}{j} \leq 1/2\}$. Note that, unlike the superiors $\SetO{i}$, the opponents $\hSetO{i}(t)$ for each arm $i$ are defined in terms of the empirical averages, and thus the algorithms know who the opponents are. 
Let the empirical divergence be 
\begin{equation*}
 I_i(t) = \sum_{j \in \hSetO{i}(t)} \myNt{i}{j} d(\hatmut{i}{j}, 1/2).
\end{equation*}
The value $\exp{(-I_i(t))}$ can be considered as the ``likelihood'' that arm $i$ is the Condorcet winner.
Let $\ist = \argmin_{i \in [K]} I_i(t)$ (ties are broken arbitrarily) and $\Ist = I_\ist(t)$.
 By definition, $\Ist \geq 0$.
RMED is inspired by the Deterministic Minimum Empirical Divergence (DMED) algorithm \citep{HondaDMED}. DMED, which is designed for  solving the standard $K$-armed bandit problem, draws arms that may be the best one with probability $\Omega(1/t)$, whereas RMED in the dueling bandit problem draws arms that are likely to be the Condorcet winner with probability $\Omega(1/t)$.
Namely, any arm $i$ that satisfies
\begin{equation}
  \EJ_i(t) = \{I_i(t) - \Ist \leq \log{t} + f(K)\}
\end{equation}
 is the candidate of the Condorcet winner and will be drawn soon. Here, $f(K)$ can be any non-negative function of $K$ that is independent of $t$. Algorithm \ref{alg:rmedbase} lists the main routine of RMED. There are several versions of RMED.
First, we introduce RMED1. 
RMED1 initially compares all pairs once (initial phase). 
Let $\Tinit = (K-1)K/2$ be the last round of the initial phase.
From $t=\Tinit+1$, it selects the arm by using a loop.
$L_C = L_C(t)$ is the set of arms in the current loop, and $L_R = L_R(t) \subset L_C(t)$ is the remaining arms of $L_C$ that have not been drawn yet in the current loop. $L_N = L_N(t)$ is the set of arms that are going to be drawn in the next loop. An arm $i$ is put into $L_N$ when it satisfies
$\{\EJ_i(t) \cap \{i \notin L_R(t)\}\}$. By definition, at least one arm (i.e. $\ist$ at the end of the current loop) is put into $L_N$ in each loop.
For arm $l(t)$ in the current loop, RMED1 selects $m(t)$ (i.e. the comparison target of $l(t)$) determined by Algorithm \ref{alg:rmedverone}. 

The following theorem, which is proven in Section \ref{sec:analysis}, describes a regret bound  of RMED1.
\begin{theorem}
For any sufficiently small $\mysmalldelta > 0$, the regret of RMED1 is bounded as:
\begin{equation*}
  \Expect[\Regret(T)] \leq \sum_{i \in \nonwinner} \frac{((1 + \mysmalldelta) \log{T} + f(K) ) \myDelta{1}{i}}{2 d(\muij{i}{1}, 1/2)} + O(K^2) + O\left(\frac{K}{\mysmalldelta^2}\right) + O(K \e^{\CoefA K-f(K)}),
\end{equation*}
where $\CoefA =  \CoefA(\{ \muij{i}{j} \}_{i,j\in[K]})$ is a constant as a function of $T$.
Therefore, by letting $\mysmalldelta = \log^{-1/3}{T}$ and choosing an $f(K) = c K^{1+\mysmallepsilon}$ for arbitrary $c,\mysmallepsilon>0$, we obtain
\begin{equation*}
  \Expect[\Regret(T)] \leq \sum_{i \in \nonwinner} \frac{ \myDelta{1}{i} \log{T}}{2 d(\muij{i}{1}, 1/2)} + O(K^{2+\mysmallepsilon}) + O(K\log^{2/3}{T}).
\end{equation*}
\label{thm:verone}
\vspace{-2em}
\end{theorem}

\vspace{-0.5em}
\subsection{Gap between the constant factor of RMED1 and the lower bound}
\vspace{-0.3em}

From the lower bound of Theorem \ref{thm:regretlower},
the $O(K\log T)$ regret bound of RMED1 is optimal up to a constant factor.
Moreover, the constant factor matches the regret lower bound of Theorem \ref{thm:regretlower}
if $\bsi=1$ for all $i \in \nonwinner$ where
\begin{align}
\bsi=\argmin_{j \in \SetO{i}}\frac{\myDelta{1}{i} + \myDelta{1}{j}}{d(\muij{i}{j}, 1/2)}.
\label{ineq:mostsuperior}
\end{align}
Here we define $d^+(p,q) = d(p,q)$ if $p<q$ and $0$ otherwise, and $x/0 = +\infty$.
Note that, there can be ties that minimize the RHS of \eqref{ineq:mostsuperior}. In that case, we may choose any of the ties as $\bsi$ to eliminate arm $i$.
For ease of explanation, we henceforth will assume that $\bsi$ is unique, but our results can be easily extended to the case of ties. 

We claim that $\bsi=1$ holds in many cases for the following
mathematical and practical reasons.
(i)
The regret of drawing a pair $(i,j),\,j\neq 1,$ is $(\myDelta{1}{i}+\myDelta{1}{j})/2$, 
whereas it is simply $\myDelta{1}{i}/2$ for the pair $(i,1)$.
Thus, $d^+(\muij{i}{j},1/2)$ has to be much larger than $d^+(\muij{i}{1},1/2)$ in order to satisfy $\bsi =j$.
 (ii) The Condorcet winner usually wins over the other arms by a large margin,
and therefore, $d^+(\muij{i}{1},1/2) \ge d^+(\muij{i}{j},1/2)$.
 For example, in the preference matrix of Example $1$ (Table \ref{tbl:example1}), $\bsiarg{3} = 1$ as long as $q <  0.79$.
Example $2$ (Table \ref{tbl:example2}) is a preference matrix based on six retrieval functions in the full-text search engine of ArXiv.org \citep{DBLP:conf/icml/YueJ11}\footnote{In the original preference matrix of \cite{DBLP:conf/icml/YueJ11}, $\muij{2}{4} \neq 1 - \muij{4}{2}$. To satisfy $\muij{2}{4} = 1 - \muij{4}{2}$, we replaced $\muij{2}{4}$ and $\muij{4}{2}$ of the original with $(\muij{2}{4} - \muij{4}{2} + 1) / 2$ and $(\muij{4}{2} - \muij{2}{4} + 1) / 2$, respectively.}. In Example $2$, $\bsi=1$ holds for all $i$, even though $\muij{1}{4} < \muij{2}{4}$.
In the case of a $16$-ranker evaluation based on the Microsoft Learning to Rank dataset (details are given in Section \ref{sec:experiment}), occasionally $\bsi \neq 1$ occurs, but the difference between the regrets of drawing arm $1$ and $\bsi$ is fairly small (smaller than $1.2$\% on average).
Nevertheless, there are some cases in which comparing arm $i$ with $1$ is not such a clever idea. 
Example $3$ (Table \ref{tbl:example3}) is a toy example in which comparing arm $i$ with $\bsi \neq 1$ makes a large difference. In Example $3$, it is clearly better to draw pairs ($2$, $4$), ($3$, $2$) and ($4$, $3$)
to eliminate arms $2$, $3$, and $4$, respectively.
Accordingly, it is still interesting to consider an algorithm that reduces regret by comparing arm $i$ with $\bsi$.

\begin{table}[h]
\begin{small}
 \label{tbl:examples}
\caption{Three preference matrices. In each example, the value at row $i$, column $j$ is $\muij{i}{j}$.}
\begin{center}
\subtable[Example $1$][caption]{
  \begin{tabular}{|l|l|l|l|}  \hline
      & 1 & 2 & 3  \\ \hline  
    1 & 0.5 & 0.7 & 0.7 \\ \hline  
    2 & 0.3 & 0.5 & \hspace{0.4em}$q$ \\ \hline  
    3 & 0.3 & 1-$q$ & 0.5 \\ \hline  
  \end{tabular}
 \label{tbl:example1}
}
\subtable[Example $2$][caption]{
 \begin{tabular}{|l|l|l|l|l|l|l|}  \hline
      & 1 & 2 & 3 & 4 & 5 & 6 \\ \hline  
    1 & 0.50 & 0.55 & 0.55 & 0.54 & 0.61 & 0.61 \\ \hline  
    2 & 0.45 & 0.50 & 0.55 & 0.55 & 0.58 & 0.60 \\ \hline  
    3 & 0.45 & 0.45 & 0.50 & 0.54 & 0.51 & 0.56 \\ \hline  
    4 & 0.46 & 0.45 & 0.46 & 0.50 & 0.54 & 0.50\\ \hline  
    5 & 0.39 & 0.42 & 0.49 & 0.46 & 0.50 & 0.51 \\ \hline  
    6 & 0.39 & 0.40 & 0.44 & 0.50 & 0.49 & 0.50 \\ \hline  
  \end{tabular}
 \label{tbl:example2}
}
\subtable[Example $3$][caption]{
  \begin{tabular}{|l|l|l|l|l|}  \hline
      & 1 & 2 & 3 & 4 \\ \hline  
    1 & 0.5 & 0.6 & 0.6 & 0.6 \\ \hline  
    2 & 0.4 & 0.5 & 0.9 & 0.1 \\ \hline  
    3 & 0.4 & 0.1 & 0.5 & 0.9 \\ \hline  
    4 & 0.4 & 0.9 & 0.1 & 0.5 \\ \hline  
  \end{tabular}
 \label{tbl:example3}
}
\end{center}
\end{small}
\end{table}

\vspace{-3em}
\subsection{RMED2 Algorithm}
\label{sec_rmed2}

\begin{algorithm}[t]
 \caption{Subroutine for selecting $m(t)$ in RMED2 and RMED2FH}
 \label{alg:rmedverttwo}
\begin{algorithmic}[1]
  \IF{RMED2}
    \STATE Update $\hatbs{l(t)}$ by \eqref{ineq:hatbsi}.
  \ENDIF
  \STATE $\hSetO{l(t)}(t) \leftarrow \{j \in [K]\setminus\{l(t)\}|  \hatmut{l(t)}{j} \leq 1/2\}$. 
  \IF{$\hatbs{l(t)} \in \hSetO{l(t)}(t)$ and $\begin{cases}
    \myNt{l(t)}{\ist} \ge \myNt{l(t)}{\hatbs{l(t)}} / \log{\log{t}} & \text{(RMED2)} \\
    \myNt{l(t)}{\ist} \ge \myNt{l(t)}{\hatbs{l(t)}} / \log{\log{T}} & \text{(RMED2FH)} \\
    \end{cases}$} \label{line:isterselect}
    \STATE $m(t) \leftarrow \hatbs{l(t)}$.
  \ELSE
    \STATE Select $m(t)$ by using Algorithm \ref{alg:rmedverone}.
  \ENDIF
\end{algorithmic}
\end{algorithm}

We here propose RMED2, which gracefully estimates $\bsi$ during a bandit game and compares arm $i$ with $\bsi$. 
RMED2 and RMED1 share the main routine (Algorithm \ref{alg:rmedbase}). The subroutine of RMED2 for selecting $m(t)$ is shown in Algorithm \ref{alg:rmedverttwo}.
Unlike RMED1, RMED2 keeps drawing pairs of arms $(i,j)$ at least $\myalpha \log{\log{t}}$ times (Line \ref{line:rmed2exploration} in Algorithm \ref{alg:rmedbase}). The regret of this exploration is insignificant since $O(\log{\log{T}}) = o(\log{T})$. 
Once all pairs have been explored more than $\myalpha \log{\log{t}}$ times, RMED2 goes to the main loop.
RMED2 determines $m(t)$ by using Algorithm \ref{alg:rmedverone} based on the estimate of $\bsi$ given by
\begin{equation}
  \hatbsi = \argmin_{j \in [K] \setminus \{i\}} \frac{\hatDeltaij{\ist}{i} + \hatDeltaij{\ist}{j}}{d^+(\hatmut{i}{j}, 1/2)}, \label{ineq:hatbsi}
\end{equation}
where ties are broken arbitrarily, $\hatDeltaij{i}{j} = 1/2 - \hatmun{i}{j}{}$ and we set $x/0 = +\infty$.
Intuitively, RMED2 tries to select $m(t) = \hatbsi$ for most rounds,
 and occasionally explores $\ist$ in order to reduce the regret increase when RMED2 fails to estimate the true $\bsi$ correctly.

\vspace{-0.7em} 
\subsection{RMED2FH algorithm}
\vspace{-0.3em} 

Although we believe that the regret of RMED2 is optimal, the analysis of RMED2 is a little bit complicated since it sometimes breaks the main loop and explores from time to time. 
For ease of analysis, we here propose RMED2 Fixed Horizon (RMED2FH, Algorithm \ref{alg:rmedbase} and \ref{alg:rmedverttwo}), which is a ``static'' version of RMED2.
Essentially, RMED2 and RMED2FH have the same mechanism. The differences are that (i) RMED2FH conducts an $\myalpha \log{\log{T}}$ exploration in the initial phase. After the initial phase (ii) $\hatbsi$ for each $i$ is fixed throughout the game. Note that, unlike RMED1 and RMED2, RMED2FH requires the number of rounds $T$ beforehand to conduct the initial $\myalpha \log{\log{T}}$ draws of each pair.
The following Theorem shows the regret of RMED2FH that matches the lower bound of Theorem \ref{thm:regretlower}.
\begin{theorem}
For any sufficiently small $\mysmalldelta>0$, the regret of RMED2FH is bounded as:
\begin{multline}
  \Expect[\Regret(T)] \leq \sum_{i \in \nonwinner}
  \frac{(\myDelta{1}{i} + \myDelta{1}{\bsi}) ((1 + \mysmalldelta) \log{T})}{2d(\muij{i}{\bsi}, 1/2)}
 + O(\myalpha K^2 \log{\log{T}}) + O(K \e^{\CoefA K - f(K)})
\\
 + O\left( \frac{K \log{T}}{\log{\log{T}}} \right) + O\left(\frac{K}{\mysmalldelta^2}\right)
 + O\left(Kf(K)\right)
, \label{rmedtworegretfst}
\end{multline}
where $\CoefA =  \CoefA(\{ \muij{i}{j} \}) > 0$ is a constant as a function of $T$.
By setting $\mysmalldelta = O((\log{T})^{-1/3})$ and choosing an $f(K) = c K^{1+\mysmallepsilon}$ ($c,\mysmallepsilon>0$) we obtain
\begin{equation}
  \Expect[\Regret(T)] \leq \sum_{i \in \nonwinner} \frac{(\myDelta{1}{i} + \myDelta{1}{\bsi}) \log{T}}{2d(\muij{i}{\bsi}, 1/2)}
 + O(\myalpha K^2 \log{\log{T}})
 + O\left( \frac{K  \log{T}}{\log{\log{T}}} \right)+ O\left( K^{2+\mysmallepsilon}\right).\label{rmedtworegretsnd}
\end{equation}
\label{thm:vertwo}
\end{theorem}
Note that all terms except the first one in \eqref{rmedtworegretsnd} are $o(\log{T})$.
From Theorems \ref{thm:regretlower} and \ref{thm:vertwo} we see that (i) RMED2FH is asymptotically optimal under the Condorcet assumption and (ii) the logarithmic term on the regret bound of RMED2FH cannot be improved even if the arms are totally ordered and the forecaster knows of the existence of the total order. The proof sketch of Theorem \ref{thm:vertwo} is in Section \ref{sec:analysis}.

\section{Experimental Evaluation}
\label{sec:experiment}

\begin{figure*}[t!]
\begin{center}
  \setlength{\subfigwidth}{.3\linewidth}
  \addtolength{\subfigwidth}{-.3\subfigcolsep}
  \begin{minipage}[t]{\subfigwidth}
  \centering
 \subfigure[Six rankers]{
 \includegraphics[scale=0.5]{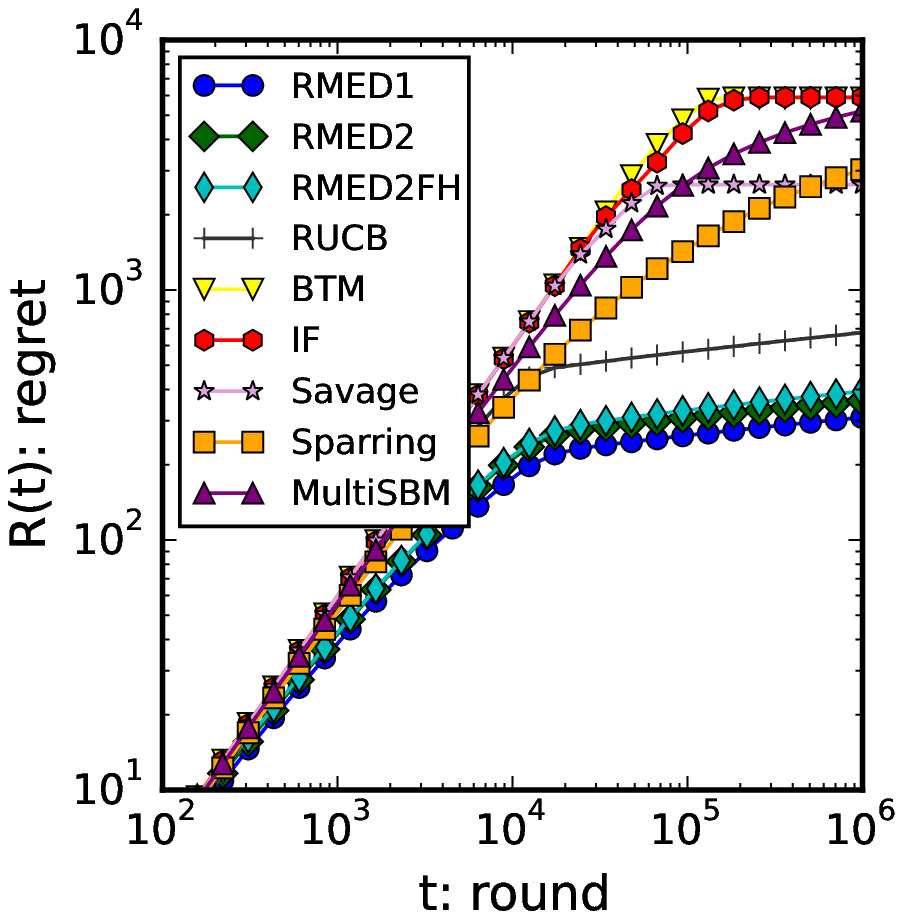}
 }
 \end{minipage}\hfill
  \begin{minipage}[t]{\subfigwidth}
  \centering
 \subfigure[Cyclic]{\includegraphics[scale=0.5]{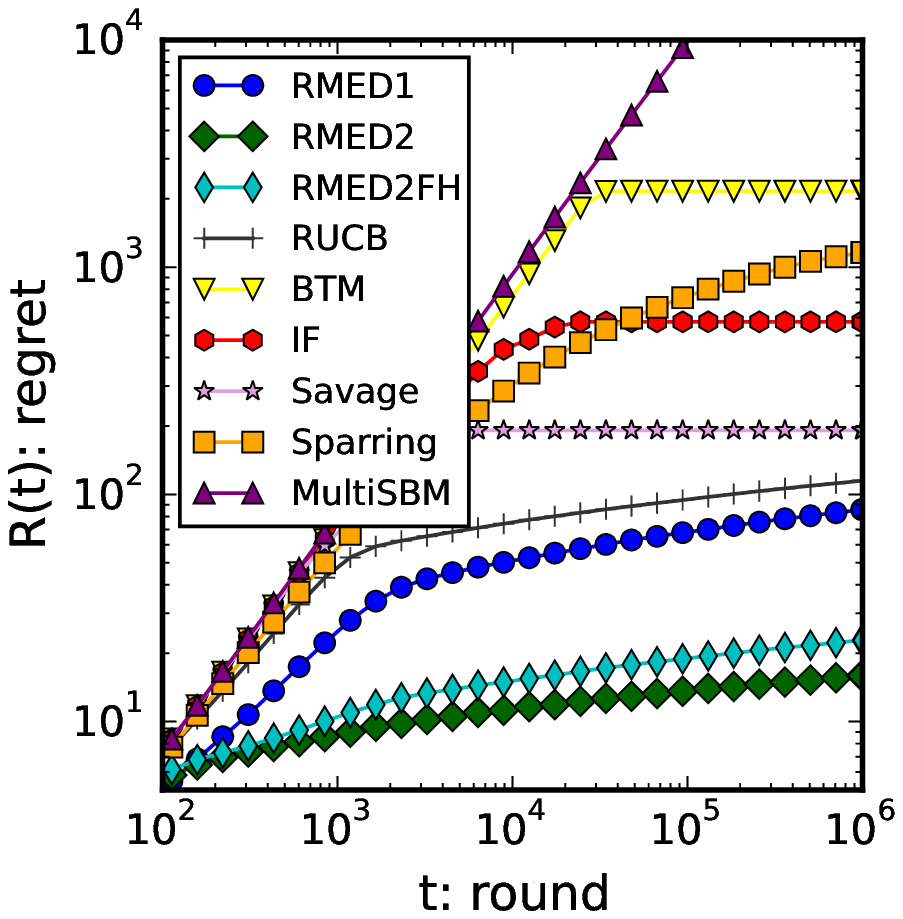}}
  \end{minipage}\hfill
  \begin{minipage}[t]{\subfigwidth}
  \centering
 \subfigure[Arithmetic]{\includegraphics[scale=0.5]{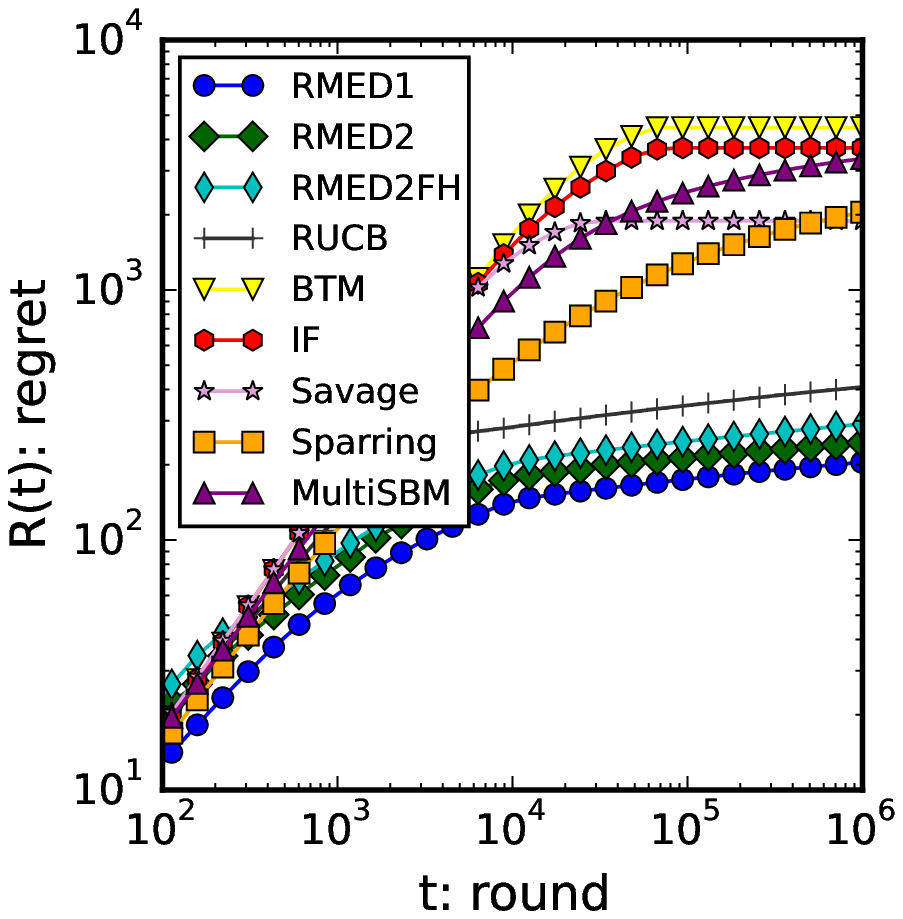}}
  \end{minipage}
  \begin{minipage}[t]{\subfigwidth}
  \centering
 \subfigure[Sushi]{\includegraphics[scale=0.5]{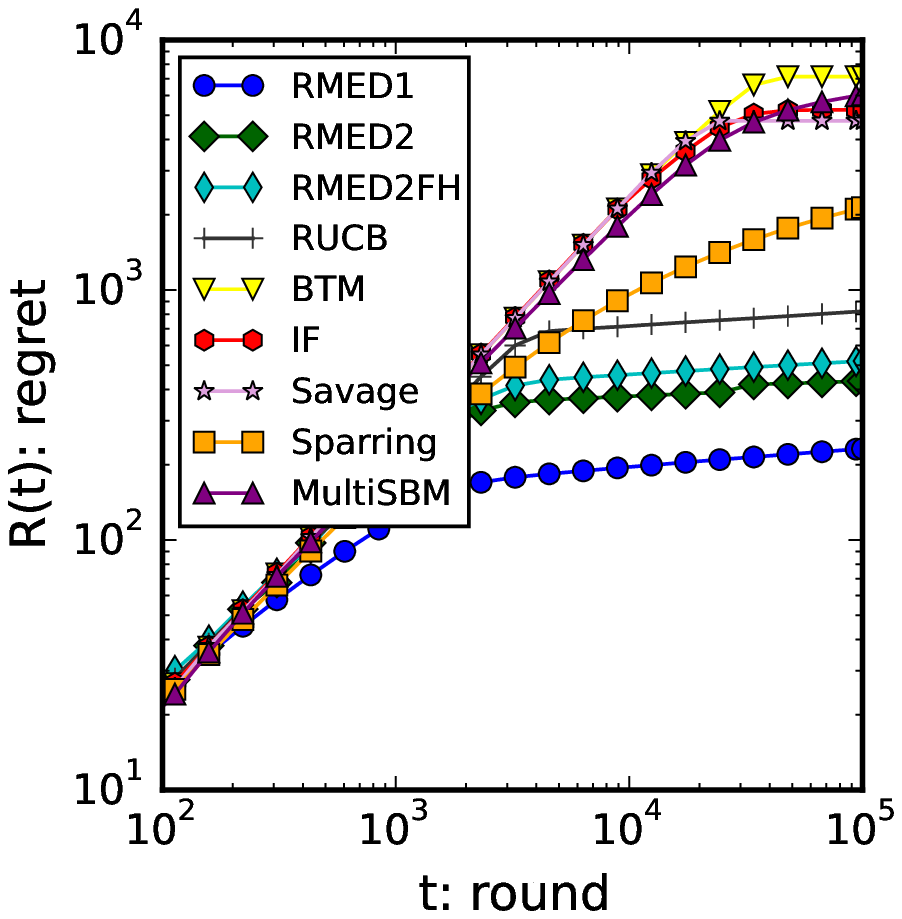}}
 \end{minipage}\hfill
  \begin{minipage}[t]{\subfigwidth}
  \centering
 \subfigure[MSLR $K=16$]{\includegraphics[scale=0.5]{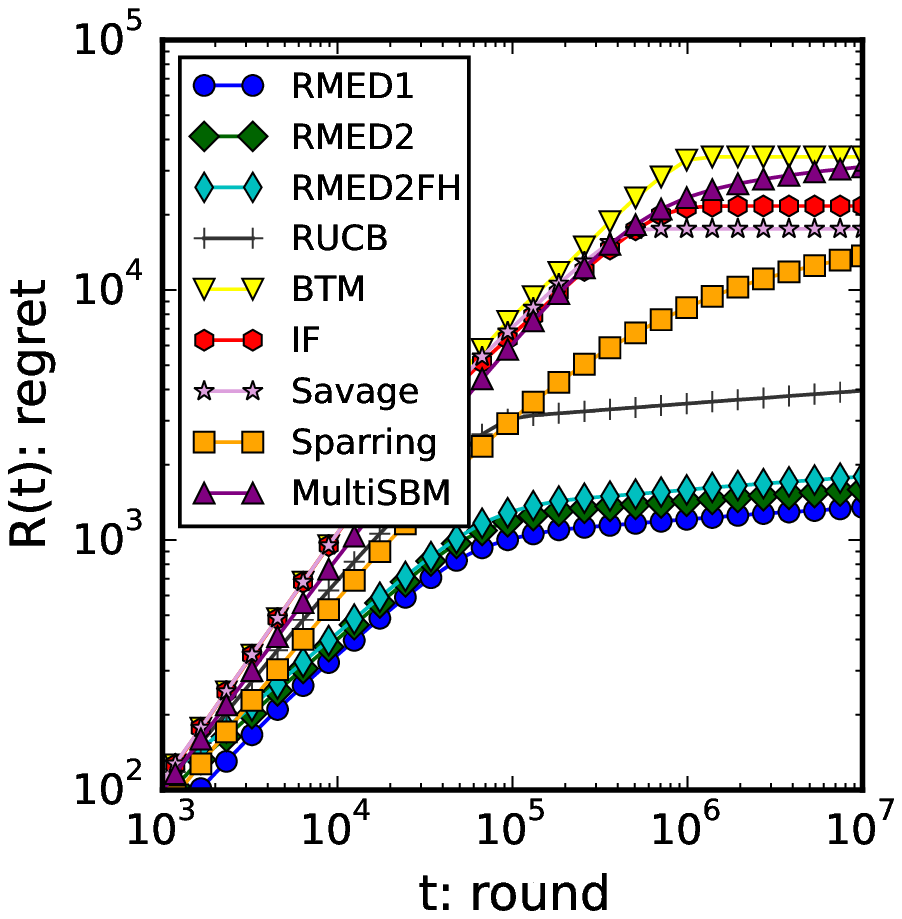}}
  \end{minipage}\hfill
  \begin{minipage}[t]{\subfigwidth}
  \centering
 \subfigure[MSLR $K=64$]{\includegraphics[scale=0.5]{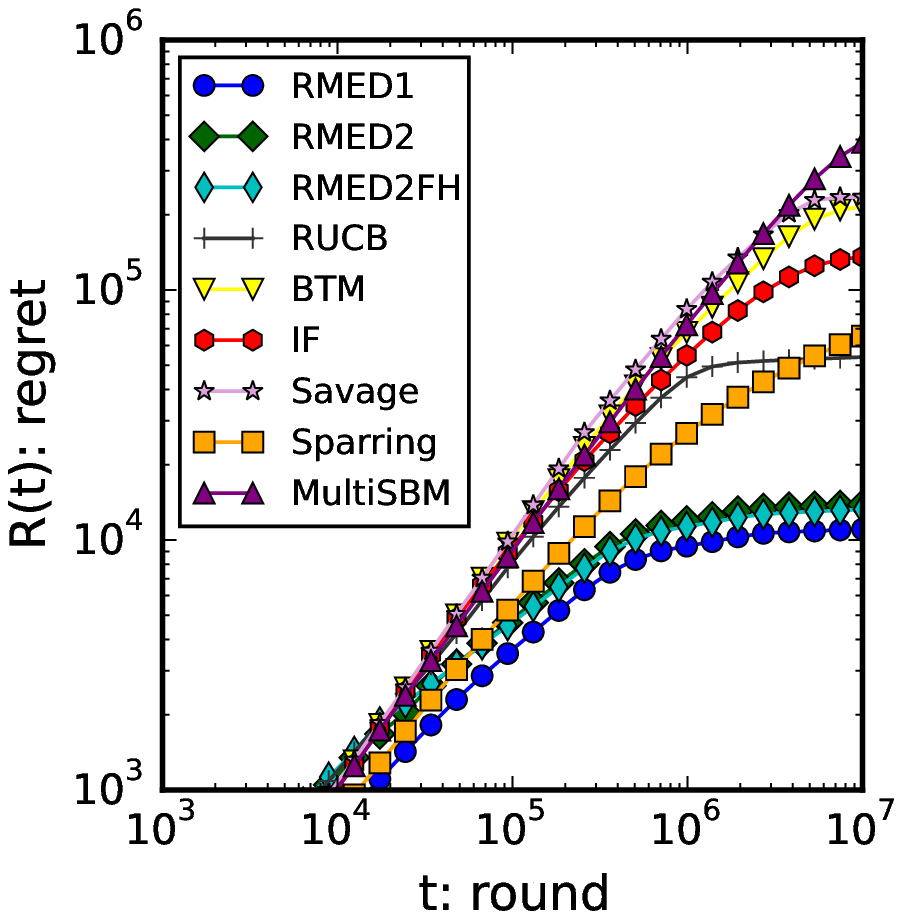}}
  \end{minipage}\hfill
\end{center}
  \caption{Regret-round log-log plots of algorithms.}
 \label{fig:regret1}
\end{figure*}%

To evaluate the empirical performance of RMED, we conducted simulations\footnote{The source code of the simulations is available at https://github.com/jkomiyama/duelingbanditlib.} with five bandit datasets (preference matrices). The datasets are as follows:

\noindent\textbf{Six rankers} is the preference matrix based on the six retrieval functions in the full-text search engine of ArXiv.org (Table \ref{tbl:example2}).

\noindent\textbf{Cyclic} is the artificial preference matrix shown in Table \ref{tbl:example3}. This matrix is designed so that the comparison of $i$ with $1$ is not optimal.

\noindent\textbf{Arithmetic} dataset involves eight arms with $\muij{i}{j} = 0.5 + 0.05 (j - i)$ and has a total order.

\noindent\textbf{Sushi} dataset is based on the Sushi preference dataset \citep{kamishimakdd2003} that contains the preferences of $5,000$ Japanese users as regards $100$ types of sushi.
We extracted the $16$ most popular types of sushi and converted them into arms with $\muij{i}{j}$ corresponding to the ratio of users who prefer sushi $i$ over $j$. 
The Condorcet winner is the mildly-fatty tuna (chu-toro). 

\noindent\textbf{MSLR:} We tested submatrices of a $136 \times 136$ preference matrix from \cite{zoghiwsdm2015}, which is derived from the Microsoft Learning to Rank (MSLR) dataset \citep{mslr2010,DBLP:journals/ir/QinLXL10} that consists of relevance information between queries and documents with more than $30$K queries. \cite{zoghiwsdm2015} created a finite set of rankers, each of which corresponds to a ranking feature in the base dataset. The value $\muij{i}{j}$ is the probability that the ranker $i$ beats ranker $j$ based on the navigational click model \citep{hofmann:tois13}. We randomly extracted $K=16, 64$ rankers in our experiments and made sub preference matrices. The probability that the Condorcet winner exists in the subset of the rankers is high (more than 90\%, c.f. Figure 1 in \cite{DBLP:conf/wsdm/ZoghiWRM14}), and we excluded the relatively small case where the Condorcet winner does not exist.

A Condorcet winner exists in all datasets.
In the experiments, the regrets of the algorithms were averaged over $1,000$ runs (Six rankers, Cyclic, Arithmetic, and Sushi), or $100$ runs (MSLR).

\subsection{Comparison among algorithms}

\begin{figure*}[t]
\begin{center}
  \setlength{\subfigwidth}{.3\linewidth}
  \addtolength{\subfigwidth}{-.3\subfigcolsep}
  \begin{minipage}[t]{\subfigwidth}
  \centering
 \subfigure[Six rankers]{
 \includegraphics[scale=0.52]{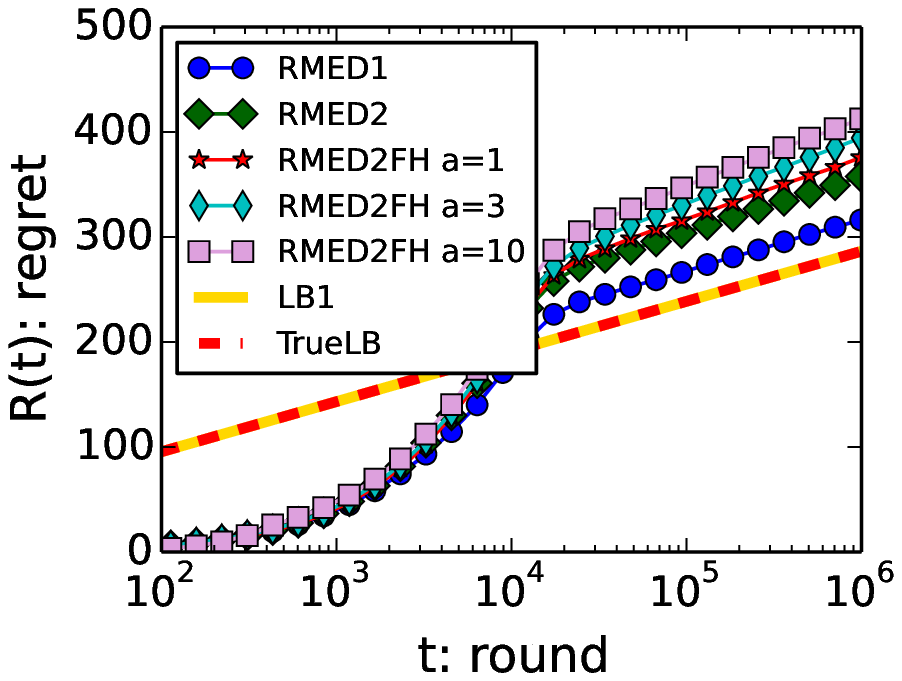}
 }
 \end{minipage}\hfill
  \begin{minipage}[t]{\subfigwidth}
  \centering
 \subfigure[Cyclic]{\includegraphics[scale=0.52]{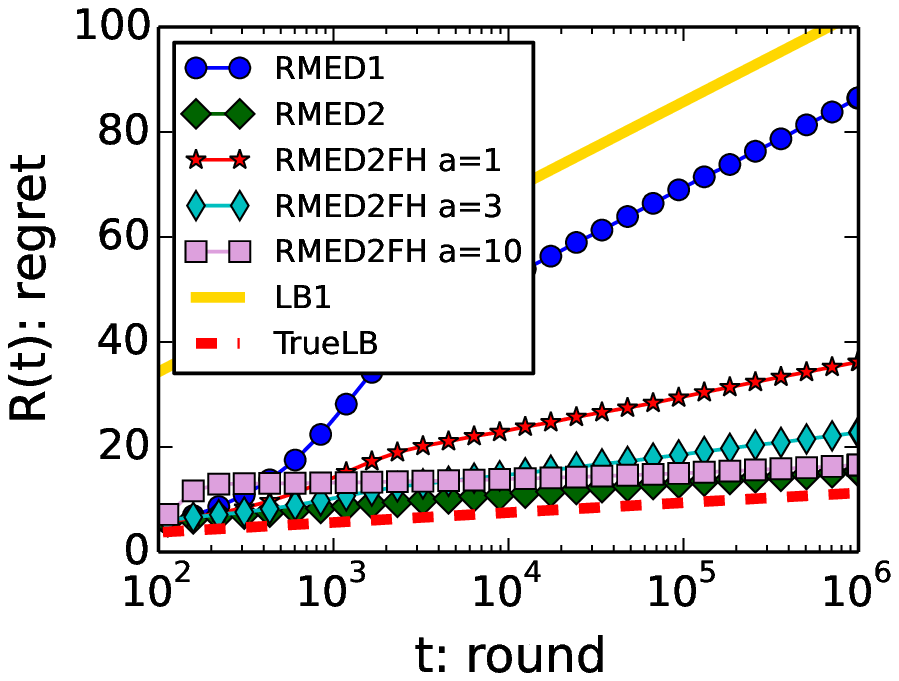}}
  \end{minipage}\hfill
  \begin{minipage}[t]{\subfigwidth}
  \centering
 \subfigure[MSLR $K=16$]{\includegraphics[scale=0.52]{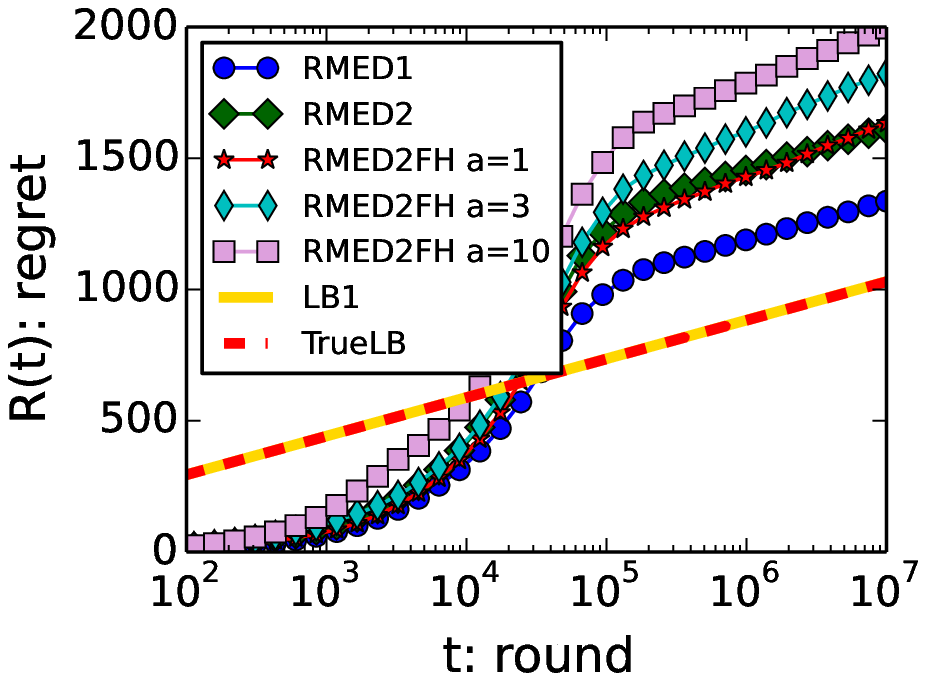}}
  \end{minipage}\hfill
\end{center}
  \caption{Regret-round semilog plots of RMED compared with theoretical bounds. We set $f(K)=0.3 K^{1.01}$ for all algorithms, and $\myalpha=3$ for RMED2.}
 \label{fig:regret_lower}
\end{figure*}%

We compared the IF, BTM with $\gamma=1.2$, RUCB with $\alpha=0.51$, Condorcet SAVAGE with $\delta=1/T$,
MultiSBM and Sparring with $\alpha=3$, and RMED algorithms. 
There are two versions of RUCB: the one that uses a randomizer in choosing $l(t)$ \citep{DBLP:conf/icml/ZoghiWMR14}, and the one that does not \citep{DBLP:journals/corr/ZoghiWMR13}. We implemented both and found that the two perform quite similarly: we show the result of the former one in this paper.
We set $f(K)=0.3 K^{1.01}$ for all RMED algorithms and set $\myalpha = 3$ for RMED2 and RMED2FH. The effect of $f(K)$ is studied in Appendix \ref{sec:depfk}.
Note that IF and BTM assume a total order among arms, which is not the case with the Cyclic, Sushi, and MSLR datasets.
MultiSBM and Sparring assume the existence of the utility of each arm, which does not allow a cyclic preference that appears in the Cyclic dataset.

Figure \ref{fig:regret1} plots the regrets of the algorithms. 
In all datasets RMED significantly outperforms RUCB, the next best excluding the different versions of RMED.
Notice that the plots are on a base $10$ log-log scale. In particular, regret of RMED1 is more than twice smaller than RUCB 
on all datasets other than Cyclic, in which RMED2 performs much better.
Among the RMED algorithms, RMED1 outperforms RMED2 and RMED2FH on all datasets except for Cyclic, in which comparing arm $i \neq 1$ with arm $1$ is inefficient. 
RMED2 outperforms RMED2FH in the five of six datasets: this could be due to the fact that RMED2FH does not update $\hatbsi$ for ease of analysis.

\subsection{RMED and asymptotic bound}

Figure \ref{fig:regret_lower} compares the regret of RMED with two asymptotic bounds. LB1 denotes the regret bound of RMED1. TrueLB is the asymptotic regret lower bound given by Theorem \ref{thm:regretlower}.

\noindent\textbf{RMED1 and RMED2:}
When $T \rightarrow \infty$, the slope of RMED1 should converge to LB1, and the ones of RMED2 and RMED2FH should converge to TrueLB. On Six rankers, LB1 is exactly the same as TrueLB, and the slope of RMED1 converges to this TrueLB. 
In Cyclic, the slope of RMED2 converges to TrueLB, whereas that of RMED1 converges to LB1, from which we see that RMED2 is actually able to estimate $\bsi \neq 1$ correctly.
In MSLR $K=16$, LB1 and TrueLB are very close (the difference is less than $1.2$\%), and RMED1 and RMED2 converge to these lower bounds. 

\noindent\textbf{RMED2FH with different values of $\myalpha$:}
We also tested RMED2FH with several values of $\myalpha$. On the one hand, with $\myalpha=1$, the initial phase of RMED2FH is too short to identify $\bsi$; as a result it performs poorly on the Cyclic dataset. On the other hand, with $\myalpha = 10$, the initial phase was too long, which incurs a practically non-negligible regret on the MSLR $K=16$ dataset.
We also tested several values of parameter $\alpha$ in RMED2FH.
We omit plots of RMED2 with $\alpha = 1,$ $10$ for the sake of readability, but we note that in our datasets the performance of RMED2 is always better than or comparable with the one of RMED2FH under the same choice of $\alpha$, although the optimality of RMED2 is not proved unlike RMED2FH.

\section{Regret Analysis}
\label{sec:analysis}

This section provides two lemmas essential for the regret analysis of RMED algorithms
and proves the asymptotic optimality of RMED1 based on these lemmas. A proof sketch on the optimal regret of RMED2FH is also given.

The crucial property of RMED is that, by constantly comparing arms with the opponents,
 the true Condorcet winner (arm $1$) actually beats all the other arms with high probability.
Let
\begin{equation*}
  \EU(t) = \bigcap_{i \in \nonwinner} \{ \hatmut{1}{i} > 1/2 \}.
\end{equation*}
Under $\EU(t)$, $\hatmut{i}{1} = 1 - \hatmut{1}{i} < 1/2$ for all $i \in \nonwinner$, and thus, $I_i(t) > 0$.
Therefore, $\EU(t)$ implies that
$\ist=\argmin_{i \in [K]} I_i(t)$ is unique with $\ist=1$
and $\Ist = I_1(t) = 0$.
Lemma \ref{lem:armoneoptimality} below shows that
the average number of rounds that $\EU^c(t)$ occurs is constant in $T$,
where the superscript $c$ denotes the complement.
\begin{lemma}
When RMED1 or RMED2FH is run, the following inequality holds:
\begin{equation}
  \Expect\left[ \sum_{t=\Tinit+1}^T \Ind\{ \EU^c(t) \} \right] = O(\e^{\CoefA K -f(K)}), \label{ineq:armoneoptimality}
\end{equation}
where $\CoefA = \CoefA(\{\muij{i}{j}\})>0$ is a constant as a function of $T$. 
\label{lem:armoneoptimality}
\end{lemma}%
Note that, since RMED2FH draws each pair $\lceil \myalpha \log{\log{T}} \rceil$ times in the initial phase, we define $\Tinit = \lceil \myalpha \log{\log{T}} \rceil (K-1)K/2$ for RMED2FH.
We give a proof of this lemma in Appendix \ref{sec:armoptimality}.
Intuitively, this lemma can be proved from the facts that arm $1$ is drawn within roughly $\e^{I_1(t)-f(K)}$ rounds
and $I_1(t)$ is not very large with high probability.

Next, for $i \in \nonwinner$ and $j \in \SetO{i}$, let
\begin{equation*}
\nsuf{i}{j} = \frac{(1+\mysmalldelta)\log{T} + f(K)}{d(\muij{i}{j}, 1/2)} + 1,
\end{equation*}
which is a sufficient number of comparisons of $i$ with $j$ to be convinced that
the arm $i$ is not the Condorcet winner.
The following lemma states that
if pair $(i,j)$ is drawn $\nsuf{i}{j}$ times then $i$ is rarely selected as $l(t)$ again.
\begin{lemma}
When RMED1 or RMED2FH is run, for $i \in \nonwinner$, $j \in \SetO{i}$,
\begin{equation*}
  \Expect\left[ \sum_{t=\Tinit+1}^T \Ind\{ l(t) = i, \myNt{i}{j} \geq \nsuf{i}{j}\} \right] = O\left(\frac{1}{\mysmalldelta^2}\right) + O(\e^{\CoefA K -f(K)}) + K.
\end{equation*}
\label{lem:suboptimalpairdrawmore}
\end{lemma} 
We prove this lemma in Appendix \ref{sec:suboptimalpairdrawmore} based on the Chernoff bound.

Now we can derive the regret bound of RMED1 based on these lemmas.

\noindent\textbf{Proof of Theorem \ref{thm:verone}:}
Since $\EU(t)$ implies
$m(t) = 1$ in RMED1,
the regret increase per round can be decomposed as:
\begin{equation}
 r(t) =  \Ind\{ \EU^c(t) \} + \sum_{i \in \nonwinner} \frac{\myDelta{1}{i}}{2} \Ind\{l(t) = i, m(t)=1, \EU(t)\}. \label{ineq:rmed1decompose}
\end{equation} 
Using Lemmas \ref{lem:armoneoptimality} and \ref{lem:suboptimalpairdrawmore}, we obtain
\begin{align*}
\lefteqn{
\Expect[\Regret(T)] \leq \Tinit + \sum_{t=\Tinit+1}^T[r(t)]
}
\nn &
 \leq \hspace{-0.2em} \frac{K(K\hspace{-0.1em}-\hspace{-0.1em}1)}{2} \hspace{-0.2em} + \hspace{-0.2em} \Expect\left[ \sum_{t=\Tinit+1}^T \hspace{-0.8em} \Ind\{ \EU^c(t) \} \right] \hspace{-0.2em}
   + \hspace{-0.8em} \sum_{i \in \nonwinner} \hspace{-0.8em}  \frac{\myDelta{1}{i}}{2} \hspace{-0.2em} \left( \hspace{-0.2em} \nsuf{i}{1} \hspace{-0.2em} + \hspace{-0.2em} \sum_{t=1}^T \Ind[ l(t) = i, m(t) = 1, \myNt{i}{1} \hspace{-0.1em} \geq \hspace{-0.1em} \nsuf{i}{1} ] \right)
\nn &
 \leq \frac{K(K-1)}{2} + O(\e^{\CoefA K -f(K)}) + \sum_{i \in \nonwinner} \frac{\myDelta{1}{i}}{2} \left( \nsuf{i}{1} + O\left(\frac{1}{\mysmalldelta^2}\right) + O(\e^{\CoefA K -f(K)}) + K \right),
\end{align*}
 which immediately completes the proof of Theorem \ref{thm:verone}. \qed
 
We also prove Theorem \ref{thm:vertwo} on the optimality of RMED2FH
based on Lemmas \ref{lem:armoneoptimality} and \ref{lem:suboptimalpairdrawmore}.
Because the full proof in Appendix \ref{subsec:proofvertwo} is a little bit lengthy, here we give its brief sketch.

\noindent\textbf{Proof sketch of Theorem \ref{thm:vertwo} (RMED2FH):}
Similar to Theorem \ref{thm:verone}, we use the fact that the $\EU^c(t)$ does not occur very often (i.e., Lemma \ref{lem:armoneoptimality}).
Under $\EU(t)$, we decompose the regret into the contributions of each arm $i \in \nonwinner$. There exists $C_2>0$ such that, for each $l(t) = i$, (i) with probability $1-O((\log{T})^{-C_2})$ RMED2FH successfully estimates $\hatbsi = \bsi$ and selects $m(t) = \bsi$ for most rounds. The optimal $O(\log{T})$ term comes from the comparison of $i$ and $\bsi$. Arm $1$ is also drawn for $O(\log{T}/\log{\log{T}}) = o(\log{T})$ times. On the other hand, (ii) with probability $O((\log{T})^{-C_2})$, RMED2FH fails to estimate $\bsi$ correctly. By occasionally comparing arm $i$ with arm $1$, we can bound the regret increase by  $O(\log{T}\log{\log{T}})$. Since $O((\log{T})^{-C_2} \times \log{T}\log{\log{T}}) = o(\log{T})$, this regret does not affect the $O(\log{T})$ factor.

\section{Discussion}
\label{sec:conclude}

We proved the regret lower bound in the dueling bandit problem.
The RMED algorithm is based on the likelihood that the arm is the Condorcet winner. RMED is proven to have the matching regret upper bound. The empirical evaluation revealed that RMED significantly outperforms the state-of-the-art algorithms. To conclude this paper, we mention three directions of future work.

First, when a Condorcet winner does not necessarily exist, the Copeland bandits \citep{DBLP:conf/icml/UrvoyCFN13} are a natural extension of our problem. Thus, seeking an effective algorithm for solving this problem will be interesting. As is well known in the field of voting theory, there are several other criteria of winners that are incompatible with the Condorcet / Copeland bandits, such as the Borda winner \citep{DBLP:conf/icml/UrvoyCFN13}. Comparing several criteria or developing an algorithm that outputs more than one of these winners should be interesting directions of future work.

Second, another direction is sequential preference elicitation problems under relative feedback that goes beyond the binary preference over pairs, such as multiscale feedback and/or preferences among three or more items.

Third, in the standard bandit problem, it is reported that KL-UCB+ \citep{lai1987,GarivierKLUCB} performs better than DMED. A study of a UCB-based optimal algorithm for the dueling bandits can yield an algorithm that outperforms RMED.

\section*{Acknowledgements}
We thank the anonymous reviewers for their useful comments. This work was supported in part by JSPS KAKENHI Grant Number 26106506.

\clearpage

\bibliographystyle{natbib}
\bibliography{bibs/manual.bib}

\begin{thebibliography}{23}
\providecommand{\natexlab}[1]{#1}
\providecommand{\url}[1]{\texttt{#1}}
\expandafter\ifx\csname urlstyle\endcsname\relax
  \providecommand{\doi}[1]{doi: #1}\else
  \providecommand{\doi}{doi: \begingroup \urlstyle{rm}\Url}\fi

\bibitem[Agrawal(1995)]{Agr95}
R.~Agrawal.
\newblock Sample mean based index policies with ${O}(\log n)$ regret for the
  multi-armed bandit problem.
\newblock \emph{Advances in Applied Probability}, 27:\penalty0 1054--1078,
  1995.

\bibitem[Ailon et~al.(2014)Ailon, Karnin, and
  Joachims]{DBLP:conf/icml/AilonKJ14}
Nir Ailon, Zohar~Shay Karnin, and Thorsten Joachims.
\newblock Reducing dueling bandits to cardinal bandits.
\newblock In \emph{ICML}, pages 856--864, 2014.

\bibitem[Auer et~al.(2002)Auer, Cesa-bianchi, and Fischer]{auerfinite}
Peter Auer, Nicol{\'o} Cesa-bianchi, and Paul Fischer.
\newblock {Finite-time Analysis of the Multiarmed Bandit Problem}.
\newblock \emph{Machine Learning}, 47:\penalty0 235--256, 2002.

\bibitem[Brochu et~al.(2010)Brochu, Brochu, and
  de~Freitas]{DBLP:conf/sca/BrochuBF10}
Eric Brochu, Tyson Brochu, and Nando de~Freitas.
\newblock A bayesian interactive optimization approach to procedural animation
  design.
\newblock In \emph{Proceedings of the 2010 Eurographics/ACM {SIGGRAPH}
  Symposium on Computer Animation, {SCA} 2010, Madrid, Spain, 2010}, pages
  103--112, 2010.

\bibitem[Bubeck(2010)]{bubeckthesis}
S{\'e}bastien Bubeck.
\newblock \emph{{Bandits Games and Clustering Foundations}}.
\newblock Theses, {Universit{\'e} des Sciences et Technologie de Lille - Lille
  I}, June 2010.

\bibitem[Garivier and Capp{\'{e}}(2011)]{GarivierKLUCB}
Aur{\'{e}}lien Garivier and Olivier Capp{\'{e}}.
\newblock The {KL-UCB} algorithm for bounded stochastic bandits and beyond.
\newblock In \emph{COLT}, pages 359--376, 2011.

\bibitem[Gemmis et~al.(2009)Gemmis, Iaquinta, Lops, Musto, Narducci, and
  Semeraro]{Gemmis09preferencelearning}
Marco~De Gemmis, Leo Iaquinta, Pasquale Lops, Cataldo Musto, Fedelucio
  Narducci, and Giovanni Semeraro.
\newblock Preference learning in recommender systems.
\newblock In \emph{In Preference Learning (PL-09) ECML/PKDD-09 Workshop}, 2009.

\bibitem[Hofmann et~al.(2013)Hofmann, Whiteson, and de~Rijke]{hofmann:tois13}
Katja Hofmann, Shimon Whiteson, and Maarten de~Rijke.
\newblock Fidelity, soundness, and efficiency of interleaved comparison
  methods.
\newblock \emph{Transactions on Information Systems}, 31(4):\penalty0 17:1--43,
  2013.

\bibitem[Honda and Takemura(2010)]{HondaDMED}
Junya Honda and Akimichi Takemura.
\newblock {An Asymptotically Optimal Bandit Algorithm for Bounded Support
  Models}.
\newblock In \emph{COLT}, pages 67--79, 2010.

\bibitem[Kamishima(2003)]{kamishimakdd2003}
Toshihiro Kamishima.
\newblock Nantonac collaborative filtering: recommendation based on order
  responses.
\newblock In \emph{KDD}, pages 583--588, 2003.

\bibitem[Lai(1987)]{lai1987}
T.~L. Lai.
\newblock Adaptive treatment allocation and the multi-armed bandit problem.
\newblock \emph{Ann. Statist.}, 15\penalty0 (3):\penalty0 1091--1114, 09 1987.

\bibitem[Lai and Robbins(1985)]{LaiRobbins1985}
T.~L. Lai and Herbert Robbins.
\newblock Asymptotically efficient adaptive allocation rules.
\newblock \emph{Advances in Applied Mathematics}, 6\penalty0 (1):\penalty0
  4--22, 1985.

\bibitem[{Microsoft Research}(2010)]{mslr2010}
{Microsoft Research}.
\newblock {Microsoft Learning to Rank Datasets}, 2010.
\newblock URL \url{http://research.microsoft.com/en-us/projects/mslr/}.

\bibitem[Qin et~al.(2010)Qin, Liu, Xu, and Li]{DBLP:journals/ir/QinLXL10}
Tao Qin, Tie{-}Yan Liu, Jun Xu, and Hang Li.
\newblock {LETOR:} {A} benchmark collection for research on learning to rank
  for information retrieval.
\newblock \emph{Inf. Retr.}, 13\penalty0 (4):\penalty0 346--374, 2010.

\bibitem[Urvoy et~al.(2013)Urvoy, Cl{\'{e}}rot, Feraud, and
  Naamane]{DBLP:conf/icml/UrvoyCFN13}
Tanguy Urvoy, Fabrice Cl{\'{e}}rot, Rapha{\"{e}}l Feraud, and Sami Naamane.
\newblock Generic exploration and k-armed voting bandits.
\newblock In \emph{ICML}, pages 91--99, 2013.

\bibitem[Yue and Joachims(2011)]{DBLP:conf/icml/YueJ11}
Yisong Yue and Thorsten Joachims.
\newblock Beat the mean bandit.
\newblock In \emph{ICML}, pages 241--248, 2011.

\bibitem[Yue et~al.(2009)Yue, Broder, Kleinberg, and
  Joachims]{DBLP:conf/colt/YueBKJ09}
Yisong Yue, Josef Broder, Robert Kleinberg, and Thorsten Joachims.
\newblock The k-armed dueling bandits problem.
\newblock In \emph{COLT}, 2009.

\bibitem[Yue et~al.(2012)Yue, Broder, Kleinberg, and
  Joachims]{DBLP:journals/jcss/YueBKJ12}
Yisong Yue, Josef Broder, Robert Kleinberg, and Thorsten Joachims.
\newblock The k-armed dueling bandits problem.
\newblock \emph{J. Comput. Syst. Sci.}, 78\penalty0 (5):\penalty0 1538--1556,
  2012.

\bibitem[Zaidan and Callison{-}Burch(2011)]{DBLP:conf/acl/ZaidanC11}
Omar Zaidan and Chris Callison{-}Burch.
\newblock Crowdsourcing translation: Professional quality from
  non-professionals.
\newblock In \emph{The 49th Annual Meeting of the Association for Computational
  Linguistics (ACL): Human Language Technologies, Proceedings of the
  Conference, 19-24 June, 2011, Portland, Oregon, {USA}}, pages 1220--1229,
  2011.

\bibitem[Zoghi et~al.(2013)Zoghi, Whiteson, Munos, and
  de~Rijke]{DBLP:journals/corr/ZoghiWMR13}
Masrour Zoghi, Shimon Whiteson, R{\'{e}}mi Munos, and Maarten de~Rijke.
\newblock Relative upper confidence bound for the k-armed dueling bandit
  problem.
\newblock \emph{CoRR}, abs/1312.3393, 2013.
\newblock URL \url{http://arxiv.org/abs/1312.3393}.

\bibitem[Zoghi et~al.(2014{\natexlab{a}})Zoghi, Whiteson, de~Rijke, and
  Munos]{DBLP:conf/wsdm/ZoghiWRM14}
Masrour Zoghi, Shimon Whiteson, Maarten de~Rijke, and R{\'{e}}mi Munos.
\newblock Relative confidence sampling for efficient on-line ranker evaluation.
\newblock In \emph{WSDM}, pages 73--82, 2014{\natexlab{a}}.

\bibitem[Zoghi et~al.(2014{\natexlab{b}})Zoghi, Whiteson, Munos, and
  de~Rijke]{DBLP:conf/icml/ZoghiWMR14}
Masrour Zoghi, Shimon Whiteson, R{\'{e}}mi Munos, and Maarten de~Rijke.
\newblock Relative upper confidence bound for the k-armed dueling bandit
  problem.
\newblock In \emph{ICML}, pages 10--18, 2014{\natexlab{b}}.

\bibitem[Zoghi et~al.(2015)Zoghi, Whiteson, and de~Rijke]{zoghiwsdm2015}
Masrour Zoghi, Shimon Whiteson, and Maarten de~Rijke.
\newblock Merge{RUCB}: A method for large-scale online ranker evaluation.
\newblock In \emph{WSDM}, 2015.

\end{thebibliography}
\clearpage

\appendix

\section{Experiment: Dependence on $f(K)$}
\label{sec:depfk}

\begin{figure}[t]
\begin{center}
\vspace{-0em}
\centerline{\includegraphics[scale=0.8]{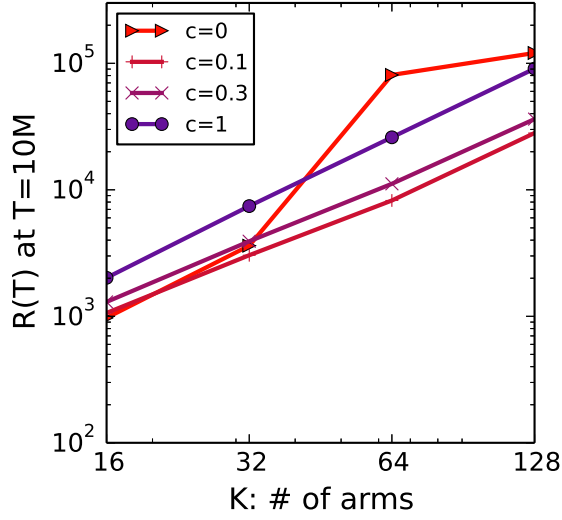}}
\vspace{-1em}
\caption{Performance of RMED1 algorithm with several values of $c$. The plot shows the regret at $T=10^7$ in the MSLR dataset with $K=16,32,64,$ and $128$.}
\label{fig:regret_vark}
\end{center}
\end{figure}

The event $\EU^c(t)$ implies a failure in identifying the Condorcet winner (i.e., $1 \neq \ist$). Although $\Expect[ \sum_{t=1}^T \EU^c(t) ] = O(\e^{\CoefA K - f(K)})$ is a constant function of $T$ for any non-negative $f(K)$, this term is not negligible with large $K$. To evaluate the effect of $f(K)$, we set $f(K) = c K^{1.01}$ and studied several values of $c$ with the MSLR dataset (Figure \ref{fig:regret_vark}). In the case of $c=0$, the regret for $K=128$ becomes 100 times that for $K=16$, which implies that the exponential dependence $O(\e^{AK})$ may not be an artifact of the proof. On the other hand, the results for $c=0.1,\,0.3$, and $1$ indicate that this term can be much improved by simply letting $c$ be a small positive value.

\section{Proofs on Regret Lower Bound}
\label{sec:lowerboundproof}

\subsection{Proof of Lemma \ref{lem:drawlower}}

\begin{mynotitleproof}{Proof of Lemma \ref{lem:drawlower}}

Let $i \in \nonwinner$ be arbitrary and $\Mat=\{\muij{i}{j}\}$ be an arbitrary preference matrix.
We consider a modified preference matrix $\Mat'$ in which the probabilities related to arm $i$ are different from $\Mat$.
Let $\SetOd{i} = \{j|j \in [K], \muij{i}{j} \le 1/2\}$, that is, $\SetOd{i} = \SetO{i} \cup \{j|j \in [K], \muij{i}{j} = 1/2\}$.
For $j \in \SetOd{i}$, $ij$ element of $\Mat'$ is $\muijd{i}{j}$ such that
\begin{equation}
  d^+(\muij{i}{j}, \muijd{i}{j}) = d(\muij{i}{j}, 1/2) + \epsilon. \label{eq:depsilondivergence}
\end{equation}
Such a $\muijd{i}{j} > 1/2$ uniquely exists for sufficiently small $\epsilon > 0$ by the monotonicity and continuity of the KL divergence. For $j \notin \SetOd{i}$, let $\muijd{i}{j} = \muij{i}{j}$.
Note that, unlike the original bandit problem, in the modified bandit problem the Condorcet winner is not arm $1$ but arm $i$. Moreover, if $\Mat \in \morder$ then $\Mat' \in \morder$.

\noindent\textbf{Notation:}
now, let $\Xij^m \in \{0,1\}$ be the result of $m$-th draw of the pair ($i$, $j$), 
\begin{equation*}
  \hKL_j(n) = \sum_{m=1}^{n} \log{\left(\frac{\Xij^m \muij{i}{j} + (1 - \Xij^m)(1-\muij{i}{j})}{\Xij^m \muijd{i}{j} + (1 - \Xij^m)(1-\muijd{i}{j})}\right)},
\end{equation*}
and $\hKL = \sum_{j \in \SetOd{i}} \hKL_j(\myNT{i}{j})$, and $\Prob'$, $\Expect'$ be the probability and the expectation with respect to the modified bandit game. Then, for any event $\EE$,
\begin{equation}
  \Prob'(\EE) = \Expect\left[\Ind\{\EE\} \exp{\left(-\hKL\right)}\right] \label{ineq:divergencediff}
\end{equation}
holds.
Let us define the events
\begin{align*}
 \ED_1 & = \left\{ \sum_{j \in \SetOd{i}} \myNT{i}{j} d(\muij{i}{j}, \muijd{i}{j}) < (1-\epsilon) \log{T},\myNT{i}{i} < \sqrt{T} \right\}, \nn
 \ED_2 & = \left\{ \hKL \leq \left(1 - \frac{\epsilon}{2}\right) \log{T} \right\}, \nn
 \ED_{12} & = \ED_1 \cap \ED_2, \nn
 \ED_{1\backslash2} & = \ED_1 \cap \ED_2^c.
\end{align*}

\noindent\textbf{First step ($\Prob\{\ED_{12}\} = o(1)$):} From \eqref{ineq:divergencediff},
\begin{align}
 \Prob'(\ED_{12})
  & \geq \Expect\left[\Ind\{\ED_{12}\} \exp{\left(- \left(1 - \frac{\epsilon}{2}\right) \log{T}\right)}\right]  = T^{-(1-\epsilon/2)} \Prob\{\ED_{12}\}.
\end{align}
By using this we have
\begin{align}
 \Prob\{\ED_{12}\}
  & \leq T^{(1-\epsilon/2)}\Prob'(\ED_{12}) \nn
  & \leq T^{(1-\epsilon/2)} \Prob'\left\{\myNT{i}{i} < \sqrt{T} \right\} \nn
  & \leq T^{(1-\epsilon/2)} \Prob'\left\{T -\myNT{i}{i} > T - \sqrt{T} \right\} \nn
  & \leq T^{(1-\epsilon/2)} \frac{ \Expect'[T -\myNT{i}{i}] }{ T -  \sqrt{T}} \text{\hspace{2em} (by the Markov inequality)}. \label{ineq:edbound}
\end{align}
Since this algorithm is strongly consistent, $\Expect'[T -\myNT{i}{i} ] \rightarrow o(T^a)$ for any $a>0$.
Therefore, the RHS of the last line of \eqref{ineq:edbound} is $o(T^{a-\epsilon/2})$, which, by choosing sufficiently small $a$, converges to zero as $T \rightarrow \infty$. In summary, $\Prob\{\ED_{12}\} = o(1)$.

\noindent\textbf{Second step ($\Prob\{\ED_{1\backslash2}\} = o(1)$):}
We have
\begin{align*}
\lefteqn{
 \Prob\{\ED_{1\backslash2}\} 
} \nn
   & = \Prob\left\{\sum_{j \in \SetOd{i}} \myNT{i}{j} d(\muij{i}{j}, \muijd{i}{j}) < (1-\epsilon) \log{T},\myNT{i}{i} < \sqrt{T}, \sum_{j \in \SetOd{i}} \hKL_j(\myNT{i}{j}) > \left(1 - \frac{\epsilon}{2}\right) \log{T} \right\} \nn
   & \leq \Prob\Biggl\{ \max_{\{n_j\} \in \Natural^{|\SetOd{i}|}, \sum_{j \in \SetOd{i}} n_j d(\muij{i}{j}, \muijd{i}{j}) < (1-\epsilon) \log{T} }\sum_{j \in \SetOd{i}} \hKL_j(n_j) > \left(1 - \frac{\epsilon}{2}\right) \log{T} \Biggr\}.
\end{align*}
Note that 
\begin{align*}
  \max_{1 \leq n \leq N} \hKL_j(n) = \max_{1 \leq n \leq N} \sum_{m=1}^n \log{\left(\frac{\Xij^{m} \muij{i}{j} + (1 - \Xij^{m})(1-\muij{i}{j})}{\Xij^{m} \muijd{i}{j} + (1 - \Xij^{m})(1-\muijd{i}{j})}\right)},
\end{align*}
 is the maximum of the sum of positive-mean random variables, and thus converges to is average  (c.f., \citealp[Lemma 10.5 in ][]{bubeckthesis}). Namely, 
\begin{equation}
 \lim_{N \rightarrow \infty} \max_{1 \leq n \leq N} \frac{\hKL_j(n)}{N} =  d(\muij{i}{j}, \muijd{i}{j}) \quad\mbox{a.s.} \label{ineq:maxsumas}
\end{equation}
Let $\delta>0$ be sufficiently small. We have,
\begin{align*}
\lefteqn{
 \frac{ \max_{\{n_j\} \in \Natural^{|\SetOd{i}|}, \sum_{j \in \SetOd{i}} n_j d(\muij{i}{j}, \muijd{i}{j}) < (1-\epsilon) \log{T} }\sum_{j \in \SetOd{i}} \hKL_j(n_j) }{\log{T}} 
} \\
 & \leq \frac{ \max_{\{n_j\} \in \Natural^{|\SetOd{i}|}, \sum_{j \in \SetOd{i}: n_j > \delta \log{T}} n_j d(\muij{i}{j}, \muijd{i}{j}) < (1-\epsilon) \log{T} }\sum_{j \in \SetOd{i}} \hKL_j(n_j) }{\log{T}} + \frac{\delta K}{\min_{j \in \SetOd{i}} d(\muij{i}{j}, \muijd{i}{j})}.
\end{align*}
Combining this with the fact that \eqref{ineq:maxsumas} holds for any $j$, we have 
\begin{equation*}
 \limsup_{N \rightarrow \infty} \frac{ \max_{\{n_j\} \in \Natural^{|\SetOd{i}|}, \sum_{j \in \SetOd{i}: n_j > \delta \log{T}} n_j d(\muij{i}{j}, \muijd{i}{j}) < (1-\epsilon) \log{T} }\sum_{j \in \SetOd{i}} \hKL_j(n_j) }{\log{T}}
 \leq 1-\epsilon \quad\mbox{a.s.,}
\end{equation*}
and thus
\begin{equation}
  \limsup_{T \rightarrow \infty} \frac{ \max_{\{n_j\} \in \Natural^{|\SetOd{i}|}, \sum_{j \in \SetOd{i}} n_j d(\muij{i}{j}, \muijd{i}{j}) < (1-\epsilon) \log{T} }\sum_{j \in \SetOd{i}} \hKL_j(n_j) }{\log{T}} \le 1-\epsilon + \Theta(\delta) \quad\mbox{a.s.}
  \label{ineq:thetadelta}
\end{equation}
By using the fact that \eqref{ineq:thetadelta} holds almost surely for any sufficiently small $\delta>0$ and $1-\epsilon/2 > 1 - \epsilon$, we have
\begin{equation*}
  \Prob\left( \max_{\{n_j\} \in \Natural^{|\SetOd{i}|}, \sum_{j \in \SetOd{i}} n_j d(\muij{i}{j}, \muijd{i}{j}) < (1-\epsilon) \log{T} }\sum_{j \in \SetOd{i}} \hKL_j(n_j) > \left(1 - \frac{\epsilon}{2}\right) \log{T} \right) = o(1).
\end{equation*}
In summary, we obtain $\Prob\left\{ \ED_{1\backslash2} \right\} = o(1)$.

\noindent\textbf{Last step:} We here have
\begin{align}
\ED_1 & 
 = \left\{ \sum_{j \in \SetOd{i}} \myNT{i}{j} d(\muij{i}{j}, \muijd{i}{j}) < (1-\epsilon) \log{T} \right\} \cap \left\{\myNT{i}{i} < \sqrt{T} \right\}
 \nn
& = \left\{ \sum_{j \in \SetOd{i}} \myNT{i}{j} (d(\muij{i}{j}, 1/2) + \epsilon) < (1-\epsilon) \log{T} \right\} \cap \left\{\myNT{i}{i} < \sqrt{T} \right\} \text{\hspace{3em} (By \eqref{eq:depsilondivergence})}
 \nn
& \supseteq \left\{ \sum_{j \in \SetOd{i}} \myNT{i}{j} (d(\muij{i}{j}, 1/2) + \epsilon) + \frac{(1-\epsilon) \log{T}}{\sqrt{T}}\myNT{i}{i} < (1-\epsilon) \log{T} \right\}, \label{ineq:probd1sup}
\end{align}
where we used the fact that $\{A<C\} \cap \{B<C\} \supseteq \{A+B<C\}$ for $A,B>0$ in the last line. 
Note that, by using the result of the previous steps, $\Prob\{\ED_1\} = \Prob\{\ED_{12}\} + \Prob\{\ED_{1\backslash2}\} = o(1)$. By using the complementary of this fact,
\begin{align*}
\Prob\left\{ \sum_{j \in \SetOd{i}} \myNT{i}{j} (d(\muij{i}{j}, 1/2) + \epsilon) + \frac{(1-\epsilon) \log{T}}{\sqrt{T}}\myNT{i}{i} \geq (1-\epsilon) \log{T} \right\} \geq \Prob\{\ED_1^c\} = 1-o(1).
\end{align*}
Using the Markov inequality yields 
\begin{equation}
 \Expect\left\{ \sum_{j \in \SetOd{i}} \myNT{i}{j} (d(\muij{i}{j}, 1/2) + \epsilon) + \frac{(1-\epsilon) \log{T}}{\sqrt{T}} \myNT{i}{i} \right\}
  \geq (1-\epsilon) (1-o(1)) \log{T}. \label{ineq:lowerlast}
\end{equation}
Because $\Expect[\myNT{i}{i}]$ is subpolynomial as a function of $T$ due to the consistency, the second term in LHS of \eqref{ineq:lowerlast} is $o(1)$ and thus negligible.
Lemma \ref{lem:drawlower} follows from the fact that \eqref{ineq:lowerlast} holds for sufficiently small $\epsilon$.
\end{mynotitleproof}

\subsection{Proof of Theorem \ref{thm:regretlower}}

\begin{mynotitleproof}{Proof of Theorem \ref{thm:regretlower}}
We have
\begin{align*}
\Regret(T)
&=
\frac{1}{2} \sum_{i \in [K]} \sum_{j \in [K] \setminus \{i\}} \frac{\myDelta{1}{i} + \myDelta{1}{j}}{2}\myNT{i}{j}
+\sum_{i \in [K]}\frac{\myDelta{1}{i} + \myDelta{1}{i}}{2}\myNT{i}{i}\nn
&\ge
\sum_{i,j \in [K]:\muij{i}{j} < 1/2}\frac{\myDelta{1}{i} + \myDelta{1}{j}}{2}\myNT{i}{j}
+\sum_{i \in [K]}\frac{\myDelta{1}{i} + \myDelta{1}{i}}{2}\myNT{i}{i}\nn
&\ge
\sum_{i\in \nonwinner}\sum_{j\in\SetO{i}}
\frac{\myDelta{1}{i} + \myDelta{1}{j}}{2}\myNT{i}{j}\nn
&=
\sum_{i\in \nonwinner}\sum_{j\in\SetO{i}}
\frac{\myDelta{1}{i} + \myDelta{1}{j}}{2d(\muij{i}{j}, 1/2)}d(\muij{i}{j}, 1/2)\myNT{i}{j}.
\end{align*}
Taking the expectation on both sides and using Lemma \ref{lem:drawlower} yield
\begin{equation*}
\Expect[\Regret(T)]\ge
\sum_{i\in \nonwinner}\min_{j\in\SetO{i}}
\frac{\myDelta{1}{i} + \myDelta{1}{j}}{2d(\muij{i}{j}, 1/2)}
(1 - o(1))\log T.
\end{equation*}
\end{mynotitleproof}

\section{Proof of Lemma \ref{lem:armoneoptimality}}
\label{sec:armoptimality}

\begin{mynotitleproof}{Proof of Lemma \ref{lem:armoneoptimality}}
This lemma essentially states that, the expected number of the rounds in which arm $1$ is underestimated is $O(1)$. We show this by bounding the expected number of rounds before arm $1$ is compared, for each fixed set of $\{\myNt{1}{s}\}$ and summing over $\{\myNt{1}{s}\}$. This technique is inspired by Lemma 16 in \cite{HondaDMED}.
Note that
\begin{align} 
\EU^c(t) 
&= 
\bigcup_{S\in2^{\nonwinner}\setminus\{\emptyset\}}
\left\{
\bigcap_{s\in S}\{ \hatmut{1}{s}\leq 1/2\}\cap
\bigcap_{s\notin S}\{ \hatmut{1}{s}>1/2\}
\right\}. \label{ineq:setsunion}
\end{align}
Now we bound the number of rounds that
the event
\begin{align*}
\bigcap_{s\in S}\{ \hatmut{1}{s}\leq 1/2\}\cap
\bigcap_{s\notin S}\{ \hatmut{1}{s}>1/2\}
\end{align*}
occurs.
Let $\Natural$ be the set of non-zero natural numbers, $n_{s}\in \Natural$ and $x_{s}\in[0,\log 2]$ be arbitrary
for each $s\in S$. Let $\hatmun{i}{j}{n}$ be the empirical estimate of $\muij{i}{j}$ at $n$-th draw of pair $(i,j)$.
If $\{\hatmun{1}{s}{n_s} \leq 1/2,\,
d^+(\hatmun{1}{s}{n_s},1/2)=x_s,\,
\myNt{1}{s}=n_s\}$ holds for $s\in S$
and $\hatmut{1}{s} > 1/2$ holds for $s\notin S$
then
\begin{align*}
I_1(t)
&=
\sum_{s\in S}n_sd^+(\hatmut{1}{s}, 1/2)
\end{align*}
and therefore $\EJ_1(t)$ holds for any
\begin{align*}
t \ge 
\exp\left(\sum_{s\in S}n_s d^+(\hatmut{1}{s},1/2) - f(K)
\right).
\end{align*}
If $\EJ_1(t)$ occurs, then arm $1$ is in $L_N$ of the next loop, and thus for some $s \in S$,  $\myNnon{1}{s}$ is incremented within $2K$ rounds. Therefore we have
\begin{multline*}
\sum_{t=\Tinit+1}^T\Ind\left[
\bigcap_{s\in S}\{ \hatmut{1}{s}\leq 1/2,\,\myNt{1}{s}=n_s\}\cap
\bigcap_{s\notin S}\{ \hatmut{1}{s}>1/2\} 
\right]
\\ \le
\exp\left(\sum_{s \in S}n_s d^+(\hatmun{1}{s}{n_s},1/2) - f(K)
\right)
+2 K.
\end{multline*}
Letting $P_s(x_s)=\Pr[\hatmun{1}{s}{n_s}\le 1/2, d^+(\hatmun{1}{s}{n_s},1/2)\ge x_s]$,
we have
\begin{align}
\lefteqn{
\Expect\left[\sum_{t=\Tinit+1}^T\Ind\left[
\bigcap_{s\in S}\{ \hatmut{1}{s}\leq 1/2,\,\myNt{1}{s}=n_s\}\cap
\bigcap_{s\notin S}\{ \hatmut{1}{s}>1/2\}
\right]
\right]
}\nn
&=
\int_{\{x_s\}\in [0,\log 2]^{|S|}}
\left(
\exp\left(\sum_{s\in S}n_s x_s - f(K)
\right)
+2K
\right)
\prod_{s\in S}\rd (-P_s(x_s))\nn
&=
\e^{-f(K)} \left\{
2K\prod_{s\in S}P_s(0)+
\prod_{s\in S}\int_{x_s\in[0,\log 2]}
\e^{n_s x_s}
\rd (-P_s(x_s))
\right\}\nn
&=
\e^{-f(K)} \left\{ 
2K\prod_{s\in S}P_s(0)+
\prod_{s\in S}
\left(
\left[-\e^{n_s x_s}P_s(x_s)\right]_0^{\log 2}
+
\int_{x_s\in[0,\log 2]}
n_s\e^{n_s x_s}
P_s(x_s)\rd x_s
\right)
\right\}
\nn &
\qquad(\mbox{integration by parts})\nn
&\le
\e^{-f(K)}\left\{
(1+2K)\prod_{s\in S}P_s(0)+
\prod_{s\in S}
\int_{x_s\in[0,\log 2]}
n_s\e^{n_s x_s}
\e^{-n_s (x_s+C_1(\muij{1}{s}, 1/2))}\rd x_s
\right\}
\nn
& \qquad(\mbox{by the Chernoff bound and Fact \ref{fact:minimumdivergencediff}, where $C_1(\mu, \mu_2) = (\mu - \mu_2)^2 / (2 \mu (1- \mu_2))$})\nn
&\le
\e^{-f(K)} \left\{
(1+2K)\prod_{s\in S}\e^{-n_s d(1/2,\muij{1}{s})}+
\prod_{s\in S}
\int_{x_s\in[0,\log 2]}
n_s\e^{-n_s C_1(\muij{1}{s}, 1/2)}\rd x_s
\right\}
\nn
&=
\e^{-f(K)} \left\{
(1+2K)\prod_{s\in S}\e^{-n_s d(1/2,\muij{1}{s})}+
\prod_{s\in S}
(\log 2)n_s
\e^{-n_s C_1(\muij{1}{s}, 1/2)}
\right\}. \label{ineq:uprobintegral}
\end{align}
By summing \eqref{ineq:uprobintegral} over $\{n_s\}$,
\begin{align*}
\lefteqn{
\sum_{t=\Tinit+1}^T
\Prob\left[
\bigcap_{s\in S}\{ \hatmut{1}{s}\leq 1/2\}\cap
\bigcap_{s\notin S}\{ \hatmut{1}{s}>1/2\}
\right]
}\nn
&\le \e^{-f(K)} \sum\dots\sum_{\hspace{-3em}\{n_s\} \in \Natural^{|S|}}\left(
(1+2K)\prod_{s\in S}\e^{-n_s d(1/2,\muij{1}{s})}+
\prod_{s\in S}
(\log 2)n_s
\e^{-n_s C_1(\muij{1}{s}, 1/2)}
\right)\nn
&\le 
\e^{-f(K)}\left\{
 (1+2K)\prod_{s\in S}\frac{1}{\e^{d(1/2,\muij{1}{s})}-1}
+(\log{2})^{|S|} \prod_{s\in S}\frac{\e^{C_1(\muij{1}{s}, 1/2)}}{(\e^{C_1(\muij{1}{s}, 1/2)}-1)^2}
\right\} 
,
\end{align*}
where we used the fact that $\sum_{n=1}^\infty\e^{-nx}=1/(\e^x+1)$ and $\sum_{n=1}^\infty n\e^{-nx}=\e^x/(\e^x+1)^2$.
Using \eqref{ineq:setsunion} and the union bound over all $S \in 2^{\nonwinner}\setminus\{\emptyset\}$, we obtain
\begin{align}
\lefteqn{
 \Expect\left[ \sum_{t=\Tinit+1}^T \Ind\{ \EU^c(t) \} \right] 
} \nn
& < \e^{-f(K)}\left\{
 (1+2K)\prod_{s \in \nonwinner}\left(1 + \frac{1}{\e^{d(1/2,\muij{1}{s})}-1}\right)
+(\log{2})^{K-1} \prod_{s \in \nonwinner}
 \left(
  1 + \frac{\e^{C_1(\muij{1}{s}, 1/2)}}{(\e^{C_1(\muij{1}{s}, 1/2)}-1)^2}
 \right)
\right\} 
\nn
& = O(\e^{\CoefA K - f(K)}), \label{ineq:euunion}
\end{align}
where $\CoefA = \log{\left\{ \max_{s \in \nonwinner} \max{\left(1 + \frac{1}{\e^{d(1/2,\muij{1}{s})}-1}, \log{2}\left(1 + \frac{\e^{C_1(\muij{1}{s}, 1/2)}}{(\e^{C_1(\muij{1}{s}, 1/2)}-1)^2}\right)\right)} \right\}}$.
\end{mynotitleproof}

\section{Proof of Lemma \ref{lem:suboptimalpairdrawmore}}
\label{sec:suboptimalpairdrawmore}

\begin{mynotitleproof}{Proof of Lemma \ref{lem:suboptimalpairdrawmore}}
Except for the first loop, arm $i$ must put into $L_N$ before $\{l(t) = i\}$. 
For $t \ge \Tinit+K+1$ (i.e., after the first loop), let $\taut{t} < t$ be the round in the previous loop in which arm $l(t)$ is put into $L_N$. In the round, $\EJ_{l(t)}(\taut{t})$ is satisfied. 
With this definition, for any two rounds $t_1, t_2 \ge \Tinit+K+1$ such that $l(t_1) = l(t_2)=i$, $t_1 \neq t_2 \Rightarrow \taut{t_1} \neq \taut{t_2}$ holds because $\taut{t_1}$ and $\taut{t_2}$ belong to different loops.
By using $\tau(t)$, we obtain
\begin{align*}
\lefteqn{
  \sum_{t=\Tinit+1}^T \Ind[l(t) = i, \myNt{i}{j} \geq \nsuf{i}{j} ]
} \nn
 & \leq  K + \hspace{-0.5em} \sum_{t=\Tinit+K+1}^T \hspace{-0.8em} \Ind[l(t) = i, \EU^c(\taut{t}) ] +  \hspace{-0.8em} \sum_{t=\Tinit+K+1}^T \hspace{-0.8em} \Ind[l(t) = i, \EU(\taut{t}), \myNt{i}{j} \geq \nsuf{i}{j} ] \nn
 & \leq K + \sum_{t=\Tinit+1}^T \Ind[\EU^c(t)] + \sum_{t=\Tinit+K+1}^T \Ind[l(t) = i, \EU(\taut{t}), \myNt{i}{j} \geq \nsuf{i}{j} ].
\end{align*}
Note that the expectation of term $\sum_{t=\Tinit+1}^T \Ind[\EU^c(t)]$ is bounded by Lemma \ref{lem:armoneoptimality}.
Between $\taut{t}$ and $t$, the only round in which pair $(i,j)$ can be compared is the round of $\{l(t) = j\}$ that occurs at most once, and thus $\myNt{i}{j} - \myNtt{i}{j}{\taut{t}} \le 1$. By using this fact, we obtain
\begin{align}
\lefteqn{
  \sum_{t=\Tinit+K+1}^T \Ind[l(t) = i, \EU(\taut{t}), \myNt{i}{j} \geq \nsuf{i}{j} ]
} \nn
 & \leq \sum_{t=\Tinit+K+1}^T \Ind[l(t) = i, \EJ_i(\taut{t}), \EU(\taut{t}), \myNtt{i}{j}{\taut{t}} \geq \nsuf{i}{j}-1 ] \nn
 & \leq \sum_{t=\Tinit+1}^T \Ind[\EJ_i(t), \EU(t), \myNtt{i}{j}{t} \geq \nsuf{i}{j}-1 ].
\end{align}
We can bound this term via $I_i(t)$ as
\begin{align}
\lefteqn{
\sum_{t=\Tinit+1}^T \Ind[\EJ_i(t), \EU(t), \myNtt{i}{j}{t} \geq \nsuf{i}{j}-1 ]
} \nn
 & \leq \sum_{n=\lceil \nsuf{i}{j} -1 \rceil}^T \Ind \left[ \bigcup_{t = \Tinit+1}^T \Bigl( I_j(t) \leq \log{t} + f(K) , \myNt{i}{j} = n \Bigr) \right] \text{\hspace{1em} (by $\EU(t) \Rightarrow I_1(t) = 0$)} \nn
 & \leq \sum_{n=\lceil \nsuf{i}{j} -1 \rceil}^T \Ind \left[ \bigcup_{t = \Tinit+1}^T \Bigl( \myNt{i}{j} = n, \myNt{i}{j} d^+(\hatmun{i}{j}{n} , 1/2) \leq \log{t} + f(K) \Bigr) \right] \nn
 & \leq \sum_{n=\lceil \nsuf{i}{j} -1 \rceil}^T \Ind \left[ (\nsuf{i}{j}-1) d^+(\hatmun{i}{j}{n}, 1/2) \leq \log{T} + f(K) \right] \nn
 & \leq \sum_{n=\lceil \nsuf{i}{j} -1 \rceil}^T \Ind \left[ d^+(\hatmun{i}{j}{n}, 1/2) \leq \frac{d(\muij{i}{j}, 1/2)}{1+\mysmalldelta} \right]. \label{ineq:epsilondivergencecomp}
\end{align}

Therefore, by letting
$\mu \in (1/2, \muij{i}{j})$ be a real number such that $d(\mu, 1/2) = \frac{d(\muij{i}{j}, 1/2)}{1+\mysmalldelta}$,
we obtain from the Chernoff bound and the monotonicity of $d^+(\cdot,1/2)$ that
\begin{align*}
 \Expect\left[ \sum_{t=\Tinit+1}^T \Ind[\EJ_i(t), \EU(t), \myNtt{i}{j}{t} \geq \nsuf{i}{j}-1 ] \right]
&\leq \sum_{n=\lceil \nsuf{i}{j}-1 \rceil}^T \Prob \left[ d^+(\hatmun{i}{j}{n}, 1/2) \leq \frac{d(\muij{i}{j}, 1/2)}{1+\mysmalldelta} \right]\nn
&\leq \sum_{n=\lceil \nsuf{i}{j}-1 \rceil}^T
\exp{\left( - d(\mu, \muij{i}{j}) n \right)}\nn
&\leq
\frac{1}{\exp{\left( d(\mu, \muij{i}{j}) \right)} - 1}  <
\frac{1}{d(\mu, \muij{i}{j})}.
\end{align*}
From the Pinsker's inequality it is easy to confirm that $d(\mu, \muij{i}{j}) = \Omega(\mysmalldelta^2)$, which completes the proof.
\end{mynotitleproof}

\section{Optimal Regret Bound: Full Proof of Theorem \ref{thm:vertwo}}

\label{subsec:proofvertwo}

\begin{mynotitleproof}{Proof of Theorem \ref{thm:vertwo}}

\noindent\textbf{Events:} 
Define
\begin{equation*}
  \EA_i = \bigcap_{i,j \in [K]} \{ |\hatmun{i}{j}{\lceil \myalpha\log{\log{T}} \rceil} - \muij{i}{j}| < \dsuf \}
\end{equation*}
for sufficiently small but fixed $\dsuf>0$.
It is easy to see from the condinuity of $d^+(\muij{i}{j},1/2)$ in $\muij{i}{j}$ that 
$\EA_i$ implies $\hatbsi = \bsi$ when we let
$\dsuf>0$ be sufficiently small with respect to
$\{\muij{i}{j}\}_{i,j \in [K]}$.
Let also 
\begin{equation*}
  \EB_i(t) = \{ \hatmut{i}{\bsi} < 1/2 \}.
\end{equation*}%
\noindent\textbf{First step (regret decomposition):} 
Like RMED1, in RMED2FH $\Expect[\EU(t)]$ holds with high probability (i.e., Lemma \ref{lem:armoneoptimality}).
In the following, we bound the regret under $\EU(t)$: let
\begin{align}
 r_i(t)
  & = \Ind\{ l(t) = i, \EU(t) \} r(t)  \nn
  & = \underbrace{ \Ind\{ l(t) = i, \EU(t), \EA_i, \EB_i(t) \} r(t) }_{\text{(A)}}
    + \underbrace{ \Ind\{ l(t) = i, \EU(t), \{\EA_i^c \cup \EB_i^c(t) \} \} r(t) }_{\text{(B)}} \label{ineq:abdecomp}
\end{align}

In the following, we first bound the terms (A) and (B), and then summarizing all terms to prove Theorem \ref{thm:vertwo}.

\noindent\textbf{Second step (bounding (A)):} 
Note that, $\{l(t) = i, \EU(t), \EA_i, \EB_i(t)\}$ is a sufficient condition for $\hatbsi = \bsi$ and $\hatbs{i} \in \hSetO{i}(t)$.
Therefore,
\begin{align*}
\lefteqn{
 \sum_{t=\Tinit+1}^T \Ind\{ l(t) = i, \EU(t), \EA_i, \EB_i(t) \} r(t) 
}\nn
 & \leq \sum_{t=\Tinit+1}^T \Ind\{ l(t) = i, \myNt{i}{\bsi} \geq \nsuf{i}{\bsi}\}  + \frac{\myDelta{1}{i} + \myDelta{1}{\bsi}}{2} \nsuf{i}{\bsi} + \frac{\myDelta{1}{i}}{2} \frac{ \nsuf{i}{\bsi}}{\log{\log{T}}}.
\end{align*}
By applying Lemma \ref{lem:suboptimalpairdrawmore} with $j = \bsi$, for sufficiently small $\mysmalldelta>0$ we have
\begin{equation*}
  \Expect\left[ \sum_{t=\Tinit+1}^T \Ind\{ l(t) = i, \myNt{i}{\bsi} \geq \nsuf{i}{\bsi}\} \right] \leq O\left(\frac{1}{\mysmalldelta^2}\right) + O(\e^{\CoefA K -f(K)}) + K. 
\end{equation*}
In summary, term (A) is bounded as:
\begin{multline}
\Expect\left[\sum_{t=\Tinit+1}^T \Ind\{ l(t) = i, \EU(t), \EA_i, \EB_i(t) \} r(t) \right] \\
\leq \frac{\myDelta{1}{i} + \myDelta{1}{\bsi}}{2} \nsuf{i}{\bsi} + O\left( \frac{ \log{T}}{\log{\log{T}}} \right) + O\left(\frac{1}{\mysmalldelta^2}\right) + O(\e^{\CoefA K -f(K)}) + K. \label{ineq:terma}
\end{multline}

\noindent\textbf{Third step (bounding (B)):} 
Now we consider the case $\{l(t) = i, \EU(t), \{\EA_i^c \cup \EB_i^c(t)\}\}$.
Under this event
$\hatbsi = \bsi$ does not always hold
but we can see that
$m(t) \in \{\hatbsi, 1\}$ still holds.
Furthermore, under this event
arm $\hatbsi$ is selected as $m(t)$ at most
$(\log{\log{T}}) N_{i,1}(T) + 1$ times
due to Line \ref{line:isterselect} of Algorithm \ref{alg:rmedverttwo}.
By using these facts, we have, 
\begin{align*}
\lefteqn{
\Expect\left[ \sum_{t=\Tinit+1}^T \Ind\{ l(t) = i, \EU(t), \{\EA_i^c \cup \EB_i^c(t) \} \} r(t) \right]
} \nn
& \leq 
\Expect\left[ \sum_{t=\Tinit+1}^T \Ind\{ l(t) = i, \EU(t), \{\EA_i^c \cup \bigcup_{t'=\Tinit+1}^T \EB_i^c(t')\} \} \right]
\nn
& \leq \Expect\left[ \sum_{t=\Tinit+1}^T  \Ind\{ l(t) = i, \myNt{i}{1} \geq \nsuf{i}{1} \} \right] \nn
 & \hspace{3em} + \Prob\left\{ \EA_i^c \cup \bigcup_{t'=\Tinit+1}^T \EB_i^c(t') \right\} \left( \nsuf{i}{1} \log{\log{T}} + 1 + \nsuf{i}{1} \right)
\nn
& \leq O\left(\frac{1}{\mysmalldelta^2}\right) + O(\e^{\CoefA K -f(K)}) + K + \Prob\left\{ \EA_i^c \cup \bigcup_{t'=\Tinit+1}^T \EB_i^c(t') \right\} O\left( \nsuf{i}{1} \log{\log{T}} \right)
 \nn & \text{\hspace{3em} (by Lemma \ref{lem:suboptimalpairdrawmore})}.
\end{align*}
The following lemma bounds $\Prob\left\{ \EA_i^c \cup \bigcup_{t'=\Tinit+1}^T \EB_i^c(t') \right\}$.
\begin{lemma}
For RMED2FH, there exists $C_2 = C_2(\{ \muij{i}{j} \}, K, \myalpha) > 0$ such that 
\begin{equation*}
  \Prob\left\{\EA_i^c \cup \bigcup_{t=\Tinit+1}^T \EB_i^c(t)\right\} = O((\log{T})^{-C_2}).
\end{equation*}
\label{lem:abcomp}
\end{lemma}%
In summary, term (B) is bounded as: 
\begin{multline}
\Expect\left[ \sum_{t=\Tinit+1} \Ind\{ l(t) = i, \EU(t), \{\EA_i^c \cup \EB_i^c(t) \} \} r(t) \right]
 \\ \leq O\left(\frac{1}{\mysmalldelta^2}\right)  + O(\e^{\CoefA K -f(K)}) + K + O\left(\nsuf{i}{1} (\log{T})^{-C_2} \log{\log{T}} \right). \label{ineq:termb}
\end{multline}

\noindent\textbf{Last step (regret bound):}
\begin{align}
\lefteqn{
\Expect[\Regret(T)]
\le \Tinit+\sum_{t=\Tinit+1}^T
 \left(\Prob\{\EU^c(t)\}+\sum_{i \in \nonwinner}\Prob\{\EU(t), l(t) = i\}r_i(t)\right)
} \nn
& \le \Tinit+\sum_{t=\Tinit+1}^T
 \left(O(\e^{\CoefA K - f(K)})+\sum_{i \in \nonwinner}\Prob(\mathrm{(A)}+\mathrm{(B)}) \right) \text{\hspace{3em}(by Lemma \ref{lem:armoneoptimality} and inequality \eqref{ineq:abdecomp})}
\nn
& \le O(\myalpha K^2 \log{\log{T}})+
O(\e^{\CoefA K - f(K)}) 
\nn &
+ \sum_{i \in \nonwinner} 
 \Biggl\{
  \frac{\myDelta{1}{i} + \myDelta{1}{\bsi}}{2} \nsuf{i}{\bsi} + O\left( \frac{\log{T}}{\log{\log{T}}} \right) \nn & \hspace{5em} + O\left(\frac{1}{\mysmalldelta^2}\right) + O(\e^{\CoefA K -f(K)}) + 2K
  + O\left(\nsuf{i}{1} (\log{T})^{-C_2} \log{\log{T}} \right)
 \Biggr\}
\nn & \text{\hspace{4em}　(by \eqref{ineq:terma} and \eqref{ineq:termb})}
\nn &
\leq O(\myalpha K^2 \log{\log{T}})+
O(K \e^{\CoefA K - f(K)}) 
+\sum_{i \in \nonwinner}
  \frac{(\myDelta{1}{i} + \myDelta{1}{\bsi}) ((1 + \mysmalldelta) \log{T})}{2d(\muij{i}{\bsi}, 1/2)} 
\nn & \hspace{2em}
  + O\left( \frac{K \log{T}}{\log{\log{T}}} \right) + O\left(\frac{K}{\mysmalldelta^2}\right)
  + O\left(K(\log{T})^{1-C_2} \log{\log{T}}\right) + O\left(Kf(K)\right). \label{rmed2_last}
\end{align}
Combining \eqref{rmed2_last} with the fact that $O\left(K(\log{T})^{1-C_2} \log{\log{T}}\right) = o\left( \frac{K \log{T}}{\log{\log{T}}} \right)$ completes the proof.
\end{mynotitleproof}

\subsection{Proof of Lemma \ref{lem:abcomp}}

\begin{mynotitleproof}{Proof of Lemma \ref{lem:abcomp}}
We bound $\Prob\{\EA_i^c\}$ and $\Prob \{ \bigcup_{t=\Tinit+1}^T \EB_i^c(t)\}$ separately.
On the one hand,
\begin{align*}
 \Prob\{\EA_i^c\} 
 & = \Prob\left\{\bigcup_{i,j \in [K]} |\hatmun{i}{j}{\lceil \myalpha\log{\log{T}} \rceil} - \muij{i}{j}| \geq \dsuf \right\} \leq \sum_{i,j \in [K]} \Prob\{ |\hatmun{i}{j}{\lceil \myalpha\log{\log{T}} \rceil} - \muij{i}{j}| \geq \dsuf \} \nn
  & \leq \sum_{i,j \in [K]} 2 \exp{(- 2 (\dsuf)^2 \myalpha\log{\log{T}})} \text{\hspace{1em} (by the Chernoff bound and Pinsker's inequality)} \nn
  & = \sum_{i,j \in [K]} 2 \left( \log{T} \right)^{- 2 (\dsuf)^2 \myalpha} = 2 K^2 \left( \log{T} \right)^{- 2 (\dsuf)^2 \myalpha} = O((\log{T})^{-C_a}),
\end{align*}
where $C_a = 2 (\dsuf)^2 \myalpha / K^2 > 0$.
On the other hand,
\begin{align*}
\lefteqn{
  \Prob \left\{ \bigcup_{t=\Tinit+1}^T \EB_i^c(t) \right\}
} \nn
   & = \Prob \left\{ \bigcup_{t=\Tinit+1}^T  \hatmut{i}{\bsi} < 1/2 \right\} 
   \leq \Prob\left( \bigcup_{n=\lceil \myalpha \log{\log{T}} \rceil}^\infty \{ \myNt{i}{\bsi}=n, \hatmun{i}{\bsi}{n} < 1/2 \} \right) \nn
   & \leq  \sum_{n= \lceil \myalpha \log{\log{T}} \rceil}^\infty \Prob\{ \myNt{i}{\bsi}=n, \hatmun{i}{\bsi}{n} < 1/2 \} \nn
   & \leq \sum_{n=\lceil \myalpha \log{\log{T}} \rceil}^\infty \exp{( -d(1/2, \muij{i}{\bsi})n )} \text{\hspace{2em} (by the Chernoff bound) }\nn
   & \leq \left(\log{T} \right)^{- \myalpha d(1/2, \muij{i}{\bsi})} \sum_{n=0}^\infty \exp{( -d(1/2, \muij{i}{\bsi})n )} 
   \nn 
   &  \leq  \left(\log{T} \right)^{- \myalpha d(1/2, \muij{i}{\bsi})} \left(1 + \frac{1}{d(1/2, \muij{i}{\bsi}) - 1} \right) = O((\log{T})^{-C_b}), 
\end{align*}
where $C_b=\myalpha d(1/2, \muij{i}{\bsi}) > 0$.
The proof is completed by letting $C_2 = \min{(C_a, C_b)}$ and taking the union bound of $\Prob\{\EA_i^c\}$ and $\Prob \{ \bigcup_{t=\Tinit+1}^T \EB_i^c(t)\}$.
\end{mynotitleproof}

\section{Facts}

\begin{fact} {\rm (The Chernoff bound)}\\
Let $X_1,\dots,X_n$ be i.i.d.\,binary random variables.
Let $\hat{X} = \frac{1}{n}\sum_{i=1}^n X_i$ and $\mu = \Expect[\hat{X}]$.
Then, for any $\epsilon > 0$,
\begin{equation*}
  \Prob( \hat{X} \geq \mu + \epsilon ) \leq \exp{\left( - d(\mu+\epsilon, \mu) n \right)}
\end{equation*}
and
\begin{equation*}
  \Prob( \hat{X} \leq \mu - \epsilon ) \leq \exp{\left( - d(\mu-\epsilon, \mu) n \right)}.
\end{equation*}
\label{fact:chernoff}
\end{fact}

\begin{fact} {\rm (The Pinsker's inequality)}\\
For $p,q \in (0,1)$, the KL divergence between two Bernoulli distributions is bounded as: 
\begin{equation*}
d(p,q) \geq 2 (p-q)^2. \label{ineq:pinsker}
\end{equation*}
\end{fact}

\begin{fact} {\rm (A minimum difference between divergences \cite[Lemma 13 in][]{HondaDMED})}\\
For any $\mu$ and $\mu_2$ satisfying $0 < \mu_2 < \mu < 1$. Let $C_1(\mu, \mu_2) = (\mu - \mu_2)^2 / (2 \mu (1- \mu_2))$. Then, for any $\mu_3 \leq \mu_2$,
\begin{equation*}
  d(\mu_3, \mu) - d(\mu_3, \mu_2) \geq C_1(\mu, \mu_2) > 0.
\end{equation*}
\label{fact:minimumdivergencediff}
\end{fact}

\end{document}